%% file: main.tex
\documentclass[conference]{IEEEtran}
\usepackage{times}

\usepackage[numbers]{natbib}
\usepackage{multicol}
\usepackage[bookmarks=true]{hyperref}
\usepackage[dvipsnames]{xcolor}
\usepackage{bm}
\usepackage{float}
\usepackage{stfloats}

\usepackage[font=small]{caption}

\input{preamble}

\input{math}
\definecolor{myblue}{rgb}{0.102, 0.263, 0.911}
\definecolor{myorange}{rgb}{0.999, 0.376, 0.208}
\definecolor{mypurple}{rgb}{0.5137, 0.098, 0.9608}

\usepackage{hyperref}
\hypersetup{
    colorlinks=true,            
}
\begin{document}

\newcommand{\method}{\textit{DemoGen}\xspace}



\title{\Huge{\textit{DemoGen:}} \huge{Synthetic Demonstration Generation\\
for Data-Efficient Visuomotor Policy Learning}}

\author{Zhengrong Xue$^{123*}$, Shuying Deng$^{1*}$, Zhenyang Chen$^{2}$, Yixuan Wang$^{1}$, Zhecheng Yuan$^{123}$, Huazhe Xu$^{123}$\vspace{0.03in}\\

$^1$Tsinghua University, $^2$Shanghai Qi Zhi Institute, $^3$Shanghai AI Lab,\quad$^*$Equal contribution\vspace{0.1in}\\

\href{https://demo-generation.github.io}{\textcolor{mypurple}{\textbf{demo-generation.github.io}}}\vspace{-0.1in}}


\twocolumn[{%
\renewcommand\twocolumn[1][]{#1}%
\maketitle
\begin{center}
    \centering
    \captionsetup{type=figure}
     \includegraphics[width=1.0\textwidth]{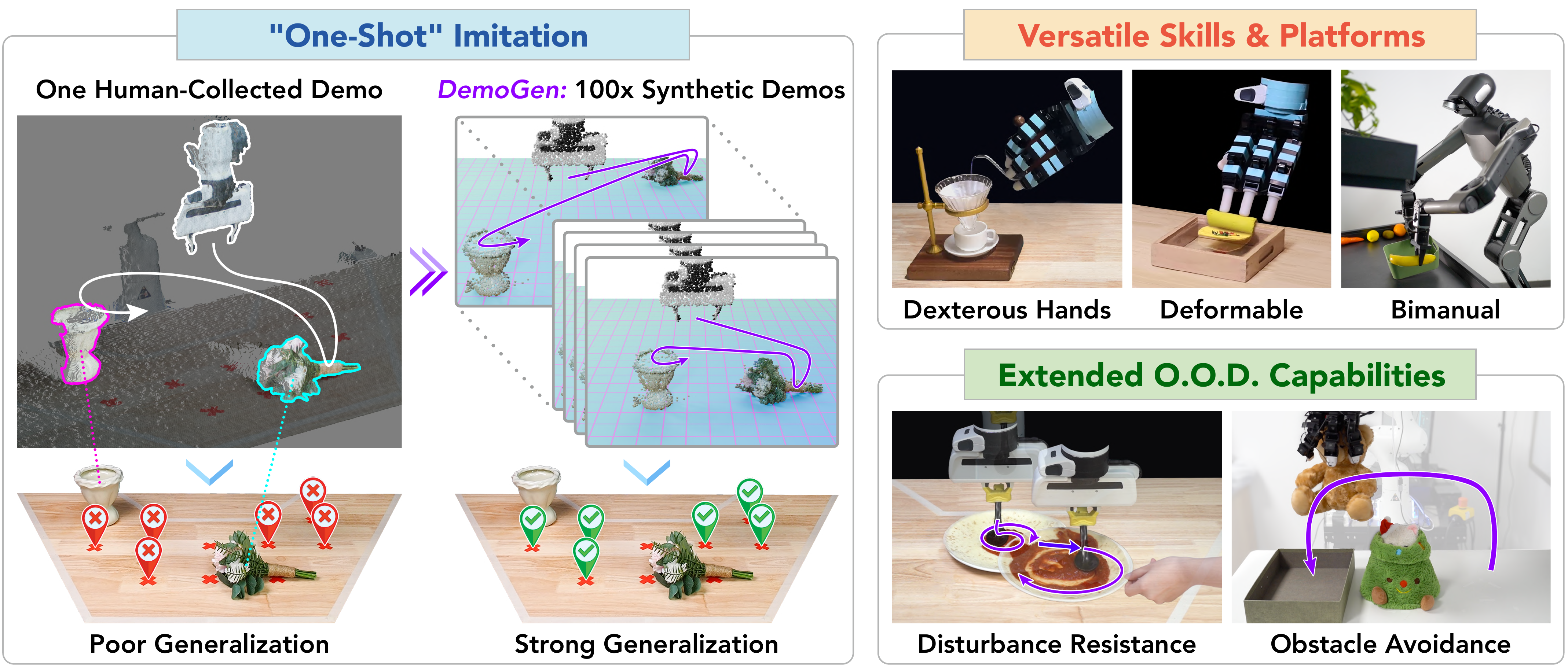}
    \caption{\textbf{\method} is a fully synthetic approach for automatic demonstration generation. \method promotes the spatial generalization ability of visuomotor policies and can facilitate one-shot imitation by adapting one human-collected demonstration into novel object configurations. 
    \method applies to various manipulation tasks and platforms and can be extended to enable additional out-of-distribution capabilities.}
    \label{fig:teaser}
\end{center}
}]

\input{sections/0_abs}

\input{sections/1_intro}
\input{sections/2_related}

\input{sections/3_empirical}

\input{sections/4_method}
\input{sections/5_sim}
\input{sections/6_real}
\input{sections/9_conclusion}


\bibliographystyle{plainnat}
\bibliography{main}

\input{sections/X_appendix}

\end{document}

%% file: preamble.tex
\usepackage{multirow}
\usepackage{multicol}
\usepackage{graphicx}
\usepackage{xspace}
\usepackage{xcolor}
\usepackage{caption}
\usepackage{wrapfig}
\usepackage{bbding}  
\usepackage{pifont}  

\usepackage{booktabs}
\usepackage{float}
\usepackage{amsmath}  
\usepackage{amssymb}  

\usepackage{colortbl}
\definecolor{ourcolor}{HTML}{99e0eb}
\definecolor{ourblue}{HTML}{27a2c3}

\definecolor{tablecolor}{HTML}{f5ecfe} 

\definecolor{tablecolor2}{HTML}{ffcdb4}
\definecolor{citecolor}{HTML}{fe7b5b}
\definecolor{grey}{rgb}{0.9, 0.9, 0.9}
\usepackage{amssymb}

\usepackage{listings}
\definecolor{gred}{rgb}{0.859,0.267,0.216}
\definecolor{ggreen}{rgb}{0.059,0.616,0.345}

\definecolor{deepblue}{HTML}{27a2c3}

\definecolor{deepred}{HTML}{fe7b5b}

\newcommand{\cc}[1]{$#1$}
\newcommand{\ccbf}[1]{\cellcolor{tablecolor}$\mathbf{#1}$}

\newcommand{\gray}[1]{\textcolor{gray}{$#1$}}

%% file: sections/0_abs.tex
\begin{abstract}
Visuomotor policies have shown great promise in robotic manipulation but often require substantial human-collected data for effective performance. 
A key factor driving the high data demands is their limited spatial generalization capability, which necessitates extensive data collection across different object configurations.
In this work, we present \method, a low-cost, fully synthetic approach for automatic demonstration generation.
Using only one human-collected demonstration per task, \method generates spatially augmented demonstrations by adapting the demonstrated action trajectory to novel object configurations. 
Visual observations are synthesized by leveraging 3D point clouds as the modality and rearranging the subjects in the scene via 3D editing.
Empirically, \method significantly enhances policy performance across a diverse range of real-world manipulation tasks, showing its applicability even in challenging scenarios involving deformable objects, dexterous hand end-effectors, and bimanual platforms. 
Furthermore, \method can be extended to enable additional out-of-distribution capabilities, including disturbance resistance and obstacle avoidance.

\end{abstract}


%% file: sections/1_intro.tex
\section{Introduction}

Visuomotor policy learning has demonstrated remarkable competence for robotic manipulation tasks~\cite{chi2023diffusion_policy,zhao2023learning,fu2024mobile,ze20243d}, yet it typically demands large volumes of human-collected data. State-of-the-art approaches often require tens to hundreds of demonstrations to achieve moderate success on complex tasks, such as spreading sauce on pizza~\cite{chi2023diffusion_policy} or making rollups with a dexterous hand~\cite{ze20243d}. More intricate, long-horizon tasks may necessitate thousands of demonstrations~\cite{zhao2024aloha}.

One key factor contributing to the data-intensive nature of these methods is their limited \textbf{spatial generalization}~\cite{saxena2024what,tan2024manibox} ability. 
Our empirical study suggests that visuomotor policies~\cite{chi2023diffusion_policy}, even when coupled with pre-trained or 3D visual encoders~\cite{nair2023r3m,radford2021learning,oquab2023dinov2,ze20243d}, exhibit limited spatial capacity, typically confined to regions adjacent to the demonstrated object configurations. 
Such limitation necessitates repeated data collection with repositioned objects until the demonstrated configurations sufficiently cover the full tabletop workspace. 
This creates a paradox: while the critical actions enabling dexterous manipulation are concentrated in a small subset of contact-rich segments, a substantial portion of human effort is spent teaching robots to approach objects in free space.

A potential solution to reduce redundant human effort is to replace the tedious relocate-and-recollect procedure with automatic demonstration generation. 
Recent advances such as MimicGen~\cite{mandlekar2023mimicgen} and its subsequent extensions~\cite{hoque2024intervengen,garrett2024skillmimicgen,jiang2024dexmimicgen} have proposed to generate demonstrations by segmenting the demonstrated trajectories based on object interactions. 
These object-centric segments are then transformed and interpolated into execution plans that fit desired spatially augmented object configurations. 
The resulting plans are then executed through open-loop rollouts on the robot, termed \textit{on-robot rollouts}, to verify their correctness and simultaneously capture the visual observations needed for policy training.

Despite their success in simulation, applying MimicGen-style strategies to real-world environments is hindered by the high costs of on-robot rollouts, which are nearly as expensive as collecting raw demonstrations. An alternative is to deploy via sim-to-real transfer~\cite{peng2018sim,torne2024reconciling,yuan2024learning}, though bridging the sim-to-real gap remains a significant challenge in robotics.

\vspace{0.2cm}   
In this work, we introduce \method, a data generation system that can be seamlessly plugged into the policy learning workflow in both simulated and physical worlds.
Recognizing the high cost of on-robot rollouts represents a major barrier to practical deployment, \method adopts a \textbf{fully synthetic} pipeline that efficiently concretizes the generated plans into spatially augmented demonstrations ready for policy training.

For action generation, \method develops the MimicGen strategy by incorporating techniques from Task and Motion Planning (TAMP)~\cite{dalal2023imitating,cheng2023nod,mandlekar2023human}, similar to the practice in the recently released SkillMimicGen~\cite{garrett2024skillmimicgen}.
Specifically, we decompose the source trajectory into \textit{motion segments} moving in free space and \textit{skill segments} involving on-object manipulation through contact. During generation, the skill segments will be transformed as a whole according to the augmented object configuration, and the motion segments will be replanned via motion planning to connect the neighboring skill segments after transformation.

With the processed actions in hand, a core challenge is obtaining spatially augmented visual observations without relying on costly on-robot rollouts. While some recent work leverages vision foundation models to manipulate the appearance of subjects and backgrounds in robotic tasks~\cite{yu2023scaling,chen2023genaug,chen2024mirage}, these techniques are not directly applicable to modifying the spatial locations of objects in an image, as 2D generative models generally lack awareness of 3D spatial relationships, such as perspective changes~\cite{xu20223d}.

\method employs a more straightforward strategy: it selects point clouds as the observation modality and synthesizes the augmented visual observations through 3D editing. 
The key insight is that point clouds, which inherently live in the 3D space, can be easily manipulated to reflect the desired spatial augmentations. 
Generating augmented point cloud observations is reduced to identifying clusters of points corresponding to the objects or robot end-effectors and then applying the same spatial transformations used in the generated action plans. 
Notably, this strategy also applies to contact-rich skill segments, as parts in contact are treated as cohesive clusters that undergo uniform transformations. 
Furthermore, the artificially applied transformations on point clouds accurately reflect the underlying physical processes, thereby minimizing the visual gap between real and synthetic observations.

\vspace{0.2cm}
Empirically, we manifest the effectiveness of \method by evaluating the performance of visuomotor policies trained on \method-generated datasets from \textbf{only one} human collected demonstration per task. To assess the impact of \method on spatial generalization, we adhere to a rigorous evaluation protocol in which the objects are placed across the entire tabletop workspace within the end-effectors' reach.

We conduct extensive real-world experiments, showing that \method can be successfully deployed on both single-arm and bi-manual platforms, using parallel-gripper and dexterous-hand end-effectors, from both third-person and egocentric observation viewpoints, and with a range of rigid-body and deformable/fluid objects. Meanwhile, the cost of generating one demonstration trajectory with \method is merely $\mathbf{0.01}$ seconds of computation. With such minimal cost, \method significantly enhances policy performance, generalizing to un-demonstrated configurations and achieving an average of $\mathbf{74.6}\%$ across $\mathbf{8}$ real-world tasks. Additionally, we demonstrate that simple extensions under the \method framework can further equip imitation learning with acquired out-of-distribution generalization capabilities such as disturbance resistance and obstacle avoidance. The code and datasets will be open-sourced to facilitate reproducibility of our results. \textbf{\textit{Please refer to the project website for robot videos.}}

%% file: sections/2_related.tex
\begin{figure*}[t]
    \centering
    \includegraphics[width=\textwidth]{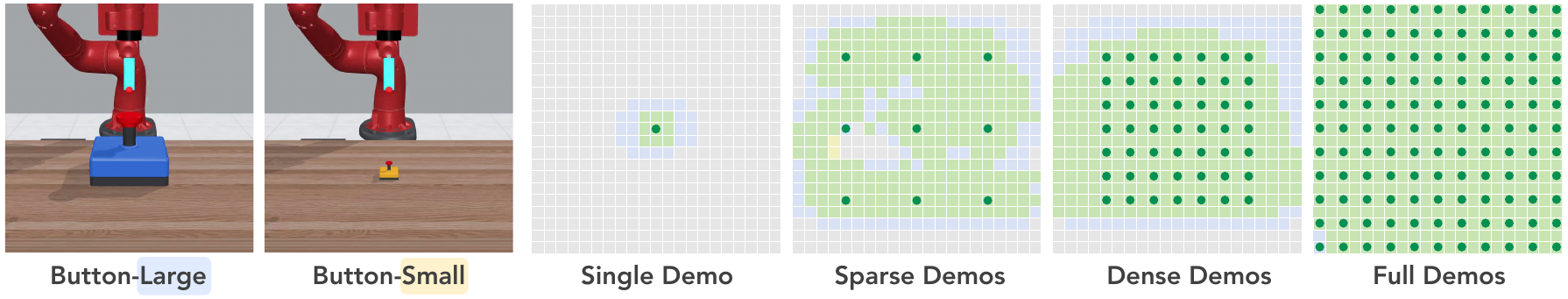}
    \caption{\textbf{Qualitative visualization of the spatial effective range.} The grid maps display discretized tabletop workspaces from a bird's-eye view under different demonstration configurations. Dark green spots mark the locations where buttons are placed during the demonstrations. Each grid cell corresponds to a policy rollout with the button placed at that location. Blue, yellow, green, and gray grids denote successful executions for the Button-Large, Button-Small, both tasks, and no tasks, respectively.}
    \label{fig:spatial_gen_vis}
    \vspace{-0.2cm}
\end{figure*}

\section{Related Works}
\subsection{Visuomotor Policy Learning}
Represented by Diffusion Policy~\cite{chi2023diffusion_policy} and its extensions~\cite{ze20243d,ke20243d,prasad2024consistency,wang2024one,wang2024equivariant}, visuomotor policy learning refers to the imitation learning methods that learn to predict actions directly from visual observations in an end-to-end fashion~\cite{levine2016end}. 
The end-to-end learning objective is a two-edged sword. Its flexibility enables visuomotor policies to learn dexterous skills from human demonstrations, extending beyond rigid-body pick-and-place. However, the absence of structured skill primitives makes such policies intrinsically data-intensive.

The conflicts between the huge data demands and the great expense of robotic data collection have driven the growing attention to data-centric research. Such efforts include more efficient data collection systems~\cite{chi2024universal,cheng2024open,li2024okami}, collaborative gathering of large-scale datasets~\cite{o2024open,khazatsky2024droid}, and empirical studies on data scaling~\cite{zhao2024aloha,lin2024data}. Instead of scaling up via pure human labor, \method aims to show that synthetic data generation can help save much of the human effort.

\subsection{Data-Efficient Imitation Learning}
Attempting to develop manipulation policies from only a handful of demonstrations, data-efficient imitation learning methods often build on the principles of Task and Motion Planning (TAMP), while incorporating imitation learning to replace some components in the TAMP pipeline.
A common approach is to learn the end-effector poses for picking and placing~\cite{zeng2021transporter,simeonov2022neural,wen2022you,xue2023useek,gao2024riemann}. The whole trajectories are generated using motion planning toolkits~\cite{kuffner2000rrt} and then executed in an open-loop manner. 
Some methods extend this idea to more complex scenarios by learning to estimate the states of manipulated objects in the environment and replaying demonstrated trajectory segments centered around the target objects~\cite{johns2021coarse,valassakis2022demonstrate,di2022learning,di2024dinobot}.
While these approaches are effective for simpler, Markovian-style tasks~\cite{vosylius2024instant}, their reliance on open-loop execution limits their application to more dexterous tasks requiring closed-loop retrying and re-planning.


In contrast, \method leverages the TAMP principles for synthetic data generation. Subsequently, the synthetic demonstrations are used to train closed-loop visuomotor policies for task resolution. In this way, \method effectively combines the merits of both approaches.

\subsection{Data Generation for Robotic Manipulation}
Automated demonstration generation offers the opportunity to breed capable visuomotor policies with significantly reduced human efforts. A branch of recent works attempts to generate demonstrations by leveraging LLM for task decomposition and then using planning or reinforcement learning for subtask resolution~\cite{wang2023gensim,hua2024gensim2,wang2023robogen}. While this paradigm enables data generation from the void, the resulting manipulation skills are often restricted by the capacity of either LLM, planning, or reinforcement learning.



An alternative line of research is exemplified by MimicGen~\cite{mandlekar2023mimicgen} and its extensions~\cite{hoque2024intervengen,garrett2024skillmimicgen,jiang2024dexmimicgen}. Unlike generating demonstrations from the void, MimicGen adapts some human-collected source demonstrations to novel object configurations by synthesizing corresponding execution plans. This approach is theoretically applicable to a wide range of manipulation skills and object types. For example, DexMimicGen~\cite{jiang2024dexmimicgen} extends MimicGen’s strategy to support bi-manual platforms equipped with dexterous hand end-effectors.
However, execution plans produced by the MimicGen framework are not ready-to-use demonstrations in the form of observation-action pairs. To bridge this gap, the MimicGen family~\cite{mandlekar2023mimicgen,hoque2024intervengen,garrett2024skillmimicgen,jiang2024dexmimicgen} relies on costly on-robot rollouts, which poses significant challenges for the deployment on physical robots.

Building upon MimicGen and its extensions, \method incorporates their strategies for generating execution plans, but replaces the expensive on-robot rollouts with an efficient, fully synthetic generation process. This enables \method to generate real-world demonstrations ready for policy training in a cost-effective manner.

%% file: sections/3_empirical.tex
\section{Empirical Study: Spatial Generalization of Visuomotor Policies}
\label{sec:empirical}


In this section, we present an empirical study examining the spatial generalization capability of visuomotor policies. We demonstrate how the lack of such generalization contributes to the data-intensive nature of learning visuomotor policies.



\subsection{Visualization of Spatial Effective Range}

Spatial generalization refers to the ability of a policy to perform tasks involving objects placed in configurations that were not seen during training. To gain an intuitive understanding of spatial generalization, we visualize the relationship between the spatial effective range of visuomotor policies and the spatial distribution of demonstration data.

\vspace{0.2cm} \noindent\textbf{Tasks.} 
We evaluate a Button-Large task adapted from the MetaWorld~\cite{yu2020metaworld} benchmark, where the robot approaches a button and presses it down. The object randomization range is modified to a $30\,\mathrm{cm} \times 40\,\mathrm{cm} = 1200\,\mathrm{cm}^2$ area on the tabletop workspace, covering most of the end-effector’s reachable space. Noticing the large size of the button makes it pressed down even if the press motion does not precisely hit the center, we also examine a more precision-demanding variant, Button-Small, where the button size is reduced by a factor of $4$. 

\vspace{0.2cm}\noindent\textbf{Policy.} 
We adopt 3D Diffusion Policy (DP3)~\cite{ze20243d} as the studied policy, as our benchmarking results indicate that 3D observations provide superior spatial generalization compared to 2D approaches. Training details are provided in Appendix~\ref{sec:appendix-policy-training}.

\vspace{0.2cm} \noindent\textbf{Evaluation.} 
To visualize the spatial effective range, we uniformly sample $21$ points along each axis within the workspace, resulting in a total of $441$ distinct button placements. Demonstrations are generated using a scripted policy, with $4$ different spatial distributions ranging from \texttt{single} to \texttt{full}. The performance of each configuration is evaluated on the $441$ placements, enabling a comprehensive assessment of spatial generalization. The visualization result is presented in Fig.~\ref{fig:spatial_gen_vis}.

\vspace{0.2cm} \noindent\textbf{Key findings.}
Overall, the spatial effective range of visuomotor policies is closely tied to the distribution of object configurations seen in the demonstrations. Specifically, the effective range can be approximated by the union of the areas surrounding the demonstrated object placements. Thus, to train a policy that generalizes well across the entire object randomization range, demonstrations must cover the full workspace, resulting in substantial data collection costs. Furthermore, as task precision requirements increase, the effective range shrinks to more localized areas, necessitating a greater number of demonstrations to adequately cover the workspace.
A more detailed analysis is available in Appendix~\ref{sec:appendix-empirical-visualize}.

\subsection{Benchmarking Spatial Generalization Capability}

The practical manifestation of the spatial generalization is reflected in the number of demonstrations required for effective policy learning. In the following benchmarking, we explore the relationship between the number of demonstrations and policy performance to determine how many demonstrations are sufficient for effective training.

\vspace{0.2cm} \noindent\textbf{Tasks.} 
To suppress the occurrence of inaccurate but successful policy rollouts, we design a Precise-Peg-Insertion task that enforces a strict fault tolerance of $1\,\mathrm{cm}$ during both the picking and insertion stages, asking for millimeter-level precision. The peg and socket are randomized within a $40\,\mathrm{cm} \times 20\,\mathrm{cm}$ area, yielding an effective workspace of $40\,\mathrm{cm} \times 40\,\mathrm{cm} = 1600\,\mathrm{cm}^2$. To examine the influence of object randomization, we also consider a \texttt{half} workspace, where the randomization range is halved for both objects, and a \texttt{fixed} setting, where object positions remain fixed. More details are listed in Appendix~\ref{sec:appendix-empirical-task}.

\vspace{0.2cm} \noindent\textbf{Policies.} In addition to Diffusion Policy (DP)~\cite{chi2023diffusion_policy} and 3D Diffusion Policy (DP3)~\cite{ze20243d} trained from scratch, we explore the potential of pre-trained visual representations to enhance spatial generalization. Specifically, we replace the train-from-scratch ResNet~\cite{he2016deep} encoder in DP with pre-trained encoders including R3M~\cite{nair2023r3m}, DINOv2~\cite{oquab2023dinov2}, and CLIP~\cite{radford2021learning}. 
Detailed implementations are provided in Appendix~\ref{sec:appendix-policy-pretrain}.

\vspace{0.2cm} \noindent\textbf{Demonstrations.}
We vary the number of demonstrations from $25$ to $400$. The object configurations are randomly sampled from a slightly larger range than the evaluation workspace to avoid performance degradation near workspace boundaries. A visualization is provided in Fig.~\ref{fig:precise-peg-insertion} in the appendix.

\begin{figure}[t]
    \includegraphics[width=0.48\textwidth]{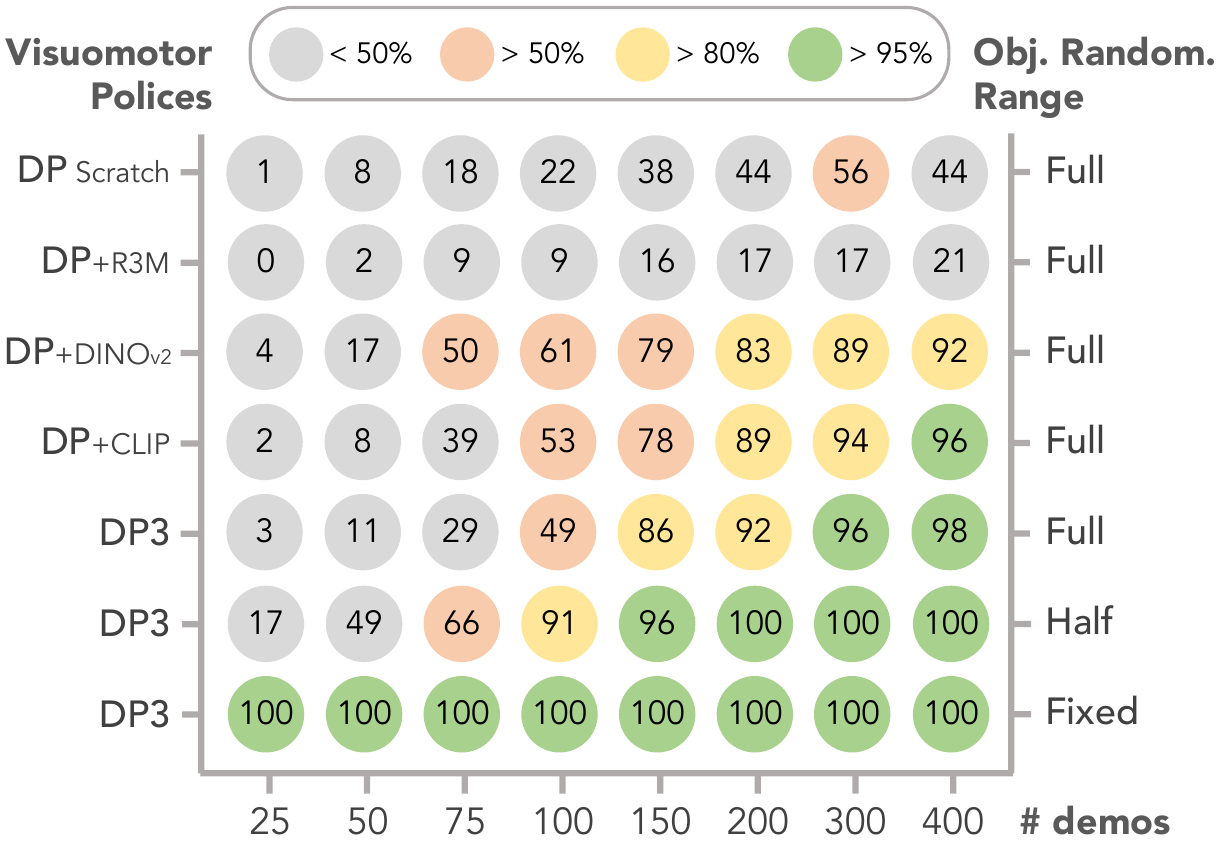}
    \caption{\textbf{Quantitative benchmarking on the spatial generalization capacity.} We report the relationship between the agent's performance in success rates and the number of demonstrations used for training when different visuomotor policies and object randomization ranges are adopted. The results are averaged over $3$ seeds.}
    \label{fig:spatial_gen_benchmark}
\end{figure}

\vspace{0.2cm} \noindent\textbf{Evaluation.}
In the \texttt{full} workspace, both the peg and socket are placed on $45$ uniformly sampled coordinates, resulting in $2025$ distinct configurations for evaluation. For the \texttt{half} and \texttt{fixed} settings, the number of evaluated configurations is $225$ and $1$, respectively. The results are presented in Fig.~\ref{fig:spatial_gen_benchmark}.

\vspace{0.2cm} \noindent\textbf{Key findings.}
The degree of object randomization significantly influences the required demonstrations. Therefore, an effective evaluation protocol for visuomotor policies must incorporate a sufficiently large workspace to provide enough object randomization. On the other hand, both 3D representations and pre-trained 2D visual encoders contribute to improved spatial generalization capabilities. However, none of these methods fundamentally resolve the spatial generalization problem. This indicates the agent’s spatial capacity is not inherently derived from the policy itself but instead develops through extensive traversal of the workspace from the given demonstrations. A more detailed analysis is provided in Appendix~\ref{sec:appendix-empirical-benchmark}.

%% file: sections/4_method.tex
\section{\method Methods}


Designed to address the conflict between the substantial data requirements of visuomotor policies and the high cost of human-collected demonstrations, \method generates spatially augmented observation-action pairs from a small set of source demonstrations. For actions, \method parses the source trajectory into object-centric motion and skill segments and applies TAMP-based adaptation. For observations, \method efficiently synthesizes the point clouds for robots and objects using a segment-and-transform strategy.

\subsection{Problem Formulation}

A visuomotor policy $\pi: \mathcal{O} \mapsto \mathcal{A}$ directly maps the visual observations $o \in \mathcal{O}$ to the predicted actions $a \in \mathcal{A}$. To train such a policy, a dataset $\mathcal{D}$ of demonstrations must be prepared. We define a source demonstration $D_{s_0} \subseteq \mathcal{D}$ as a trajectory of paired observations and actions conditioned on an initial object configuration: $D_{s_0} = (d_0, d_1, \dots, d_{L-1} | s_0)$, where each $d_t = (o_t, a_t)$ represents an observation-action pair, $s_0$ denotes the initial configuration, and $L$ is the trajectory length. 
\method is designed to augment a human-collected source demonstration by generating a new demonstration conditioned on a different initial object configuration:
\begin{equation*}
    \hat{D}_{s'_0} = (\hat{d}_0, \hat{d}_1, \dots, \hat{d}_{L-1} | s'_0).
\end{equation*}

Specifically, assuming the task involves the sequential manipulation of $K$ objects $\{O_1, O_2, \dots, O_K\}$, the initial object configuration $s_0$ is defined as the set of initial poses of these objects: $s_0 = \{\mathbf{T}_0^{O_1}, \mathbf{T}_0^{O_2}, \dots, \mathbf{T}_0^{O_K}\}$, where $\mathbf{T}^O_t$ denotes the $\mathrm{SE(3)}$ transformation from the world frame to an object $O$ at time step $t$.
The action $a_t$ consists of the robot arm and robot hand commands, represented as $a_t = (a_t^{\mathrm{arm}}, a_t^{\mathrm{hand}})$, where $a_t^{\mathrm{arm}}\triangleq \mathbf{A}^{\mathrm{EE}}_t$ is the target $\mathrm{SE(3)}$ end-effector pose in the world frame, and $a_t^{\mathrm{hand}}$ can either be a binary signal for a parallel gripper’s open/close action or a higher-dimensional vector for controlling the joints of a dexterous hand. The observation $o_t$ includes both the point cloud data and the proprioceptive feedback from the robot: $o_t = (o^{\mathrm{pcd}}_t, o^{\mathrm{arm}}_t, o^{\mathrm{hand}}_t)$, where $o^{\mathrm{arm}}_t$ and $o^{\mathrm{hand}}_t$ reflect the current state of the end-effector, with the same dimensionality as the corresponding actions.

 \begin{figure}
    \centering
    \includegraphics[width=0.9\linewidth]{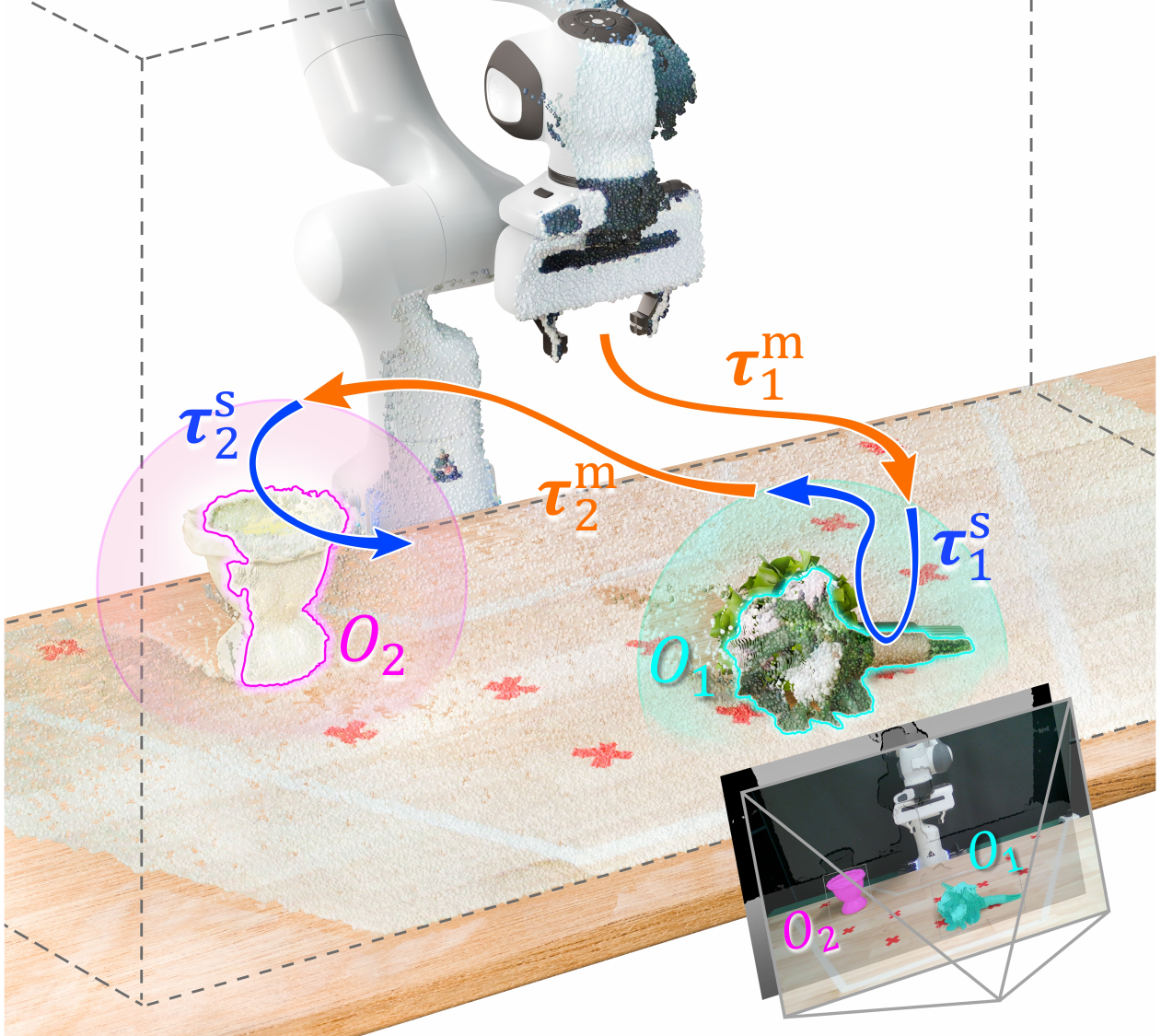}
    \caption{\textbf{Pre-processing the source demonstration.} The raw point cloud observations are processed by cropping, clustering, and down-sampling. The source action trajectory is parsed into \textcolor{myorange}{motion} and \textcolor{myblue}{skill} segments by referring to the semantic masks of manipulated objects.}
    \label{fig:method_parse}
    \vspace{-0.2cm}
\end{figure}

\subsection{Pre-processing the Source Demonstration}
\label{sec:method-preprocess}

\noindent \textbf{Segmented point cloud observations.}  
To improve the practical applicability in real-world scenarios, we utilize a single-view RGBD camera for point cloud acquisition.
The raw point cloud observations are first preprocessed by cropping the redundant points from the background and table surface. We assume the retained points are associated with either the manipulated object(s) or the robot’s end-effector. A clustering operation~\cite{ester1996density} is then applied to filter out the outlier points in noisy real-world observations. Subsequently, the point cloud is downsampled to a fixed number of points (e.g., 512 or 1024) using farthest point sampling to facilitate policy learning~\cite{qi2017pointnet}. 

For the first frame of the trajectory, we employ Grounded SAM~\cite{ren2024grounded} to obtain the segmentation masks for the manipulated objects from the RGB image. These masks are then applied to the pixel-aligned depth image and projected onto the 3D point cloud, as shown in Fig.~\ref{fig:method_parse}.

\vspace{0.2cm} 
\noindent \textbf{Parsing the source trajectory.}  
Following previous work~\cite{mandlekar2023mimicgen, garrett2024skillmimicgen}, we assume that the execution trajectory can be parsed into a sequence of object-centric segments. Noticing that the robot must initially \textcolor{myorange}{approach} the object in free space before engaging in on-object manipulation through \textcolor{myblue}{contact}, each object-centric segment can be further subdivided into two stages: \textcolor{myorange}{motion} and \textcolor{myblue}{skill}. For example, in the task illustrated in Fig.~\ref{fig:method_parse}, the trajectory is divided into four stages: \textcolor{myorange}{1) \textit{move} to the flower}, \textcolor{myblue}{2) \textit{pick} up the flower}, \textcolor{myorange}{3) \textit{transfer} the flower to the vase}, and \textcolor{myblue}{4) \textit{insert} the flower into the vase}.

We can easily identify the skill segments associated with a given object by checking whether the distance between the geometric center of the object's point cloud and the robot's end-effector falls within a predefined threshold, as illustrated by the spheres in Fig.~\ref{fig:method_parse}. The intermediate trajectories between two skill segments are classified as motion segments.

Formally, we represent an interval of time stamps as $\bm{\tau}$:
\begin{equation*}
    \bm{\tau} = (t_{\mathrm{start}},~ t_{\mathrm{start}}+1,~ \dots, ~t_{\mathrm{end}}-1,~ t_{\mathrm{end}})\subseteq(0,1,\dots,L-1),
\end{equation*}
which can be used as an \textit{index sequence} for the extraction of the corresponding segments from a sequence of demonstrations, actions, or observations. For instance, $d[\bm{\tau}] = (d_{t_{\mathrm{start}}},d_{t_{\mathrm{start}}+1},\dots,d_{t_{\mathrm{end}}-1}, d_{t_{\mathrm{end}}})$ represents the extracted subset of source demonstration indexed by $\bm{\tau}$.
Using this notation, we parse the source demonstration into alternating \textcolor{myorange}{motion} and \textcolor{myblue}{skill} segments according to the index sequence
$(\textcolor{myorange}{\bm{\tau}^{\mathrm{m}}_1}, \textcolor{myblue}{\bm{\tau}^{\mathrm{s}}_1}, \dots, \textcolor{myorange}{\bm{\tau}^{\mathrm{m}}_K}, \textcolor{myblue}{\bm{\tau}^{\mathrm{s}}_K})$:

\begin{equation*}
    D_{s_0} = (\textcolor{myorange}{d[{\bm{\tau}^{\mathrm{m}}_1}]}, \textcolor{myblue}{d[{\bm{\tau}^{\mathrm{s}}_1}]}, \dots, \textcolor{myorange}{d[{\bm{\tau}^{\mathrm{m}}_K}]}, \textcolor{myblue}{d[{\bm{\tau}^{\mathrm{s}}_K}]} | s_0).
\end{equation*}


\begin{figure}[b]
    \vspace{-0.2cm}
    \centering
    \includegraphics[width=\linewidth]{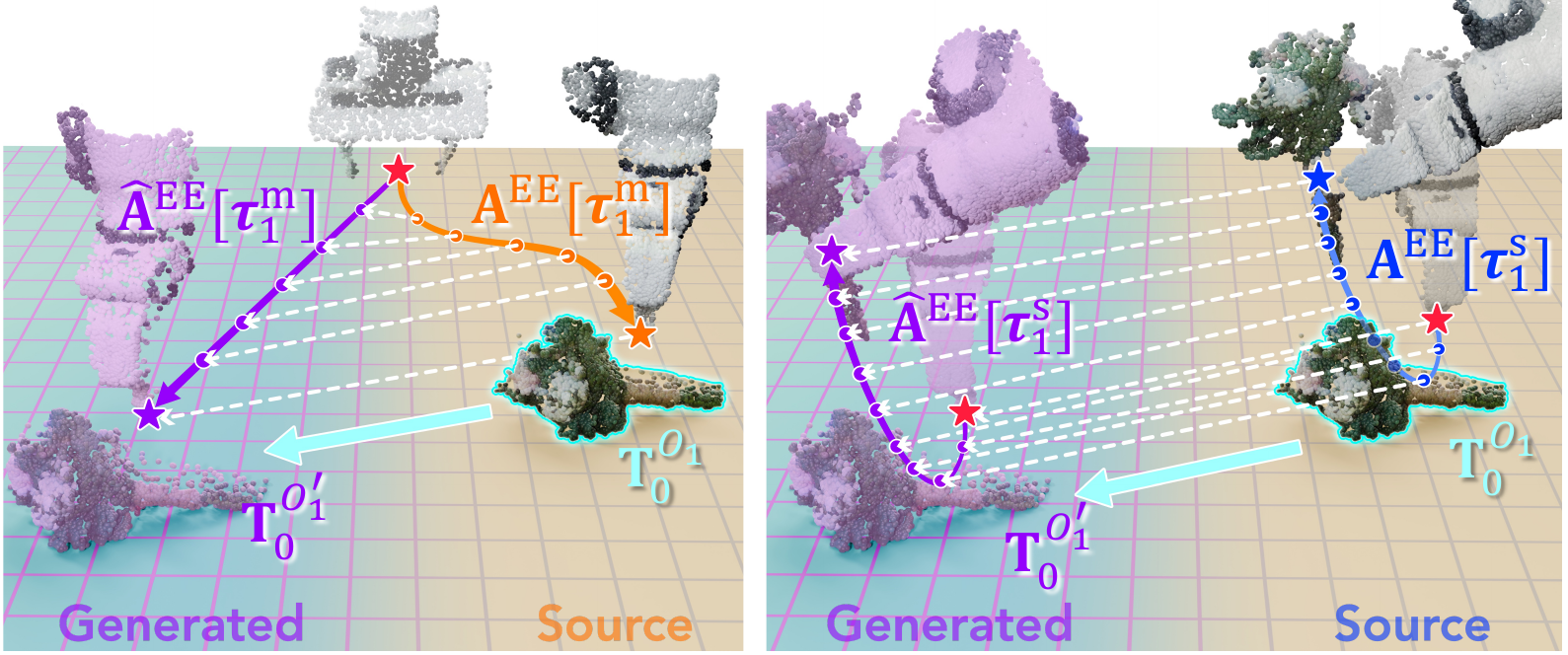}
    \caption{\textbf{Illustrations for action generation.} (Left) Actions in the \textcolor{myorange}{motion} stage are planned to connect the neighboring skill segments. (Right) Actions in the \textcolor{myblue}{skill} stage undergo a uniform transformation.}
    \label{fig:method-motion-skill}
\end{figure}

\begin{figure*}
    \centering
    \includegraphics[width=1\linewidth]{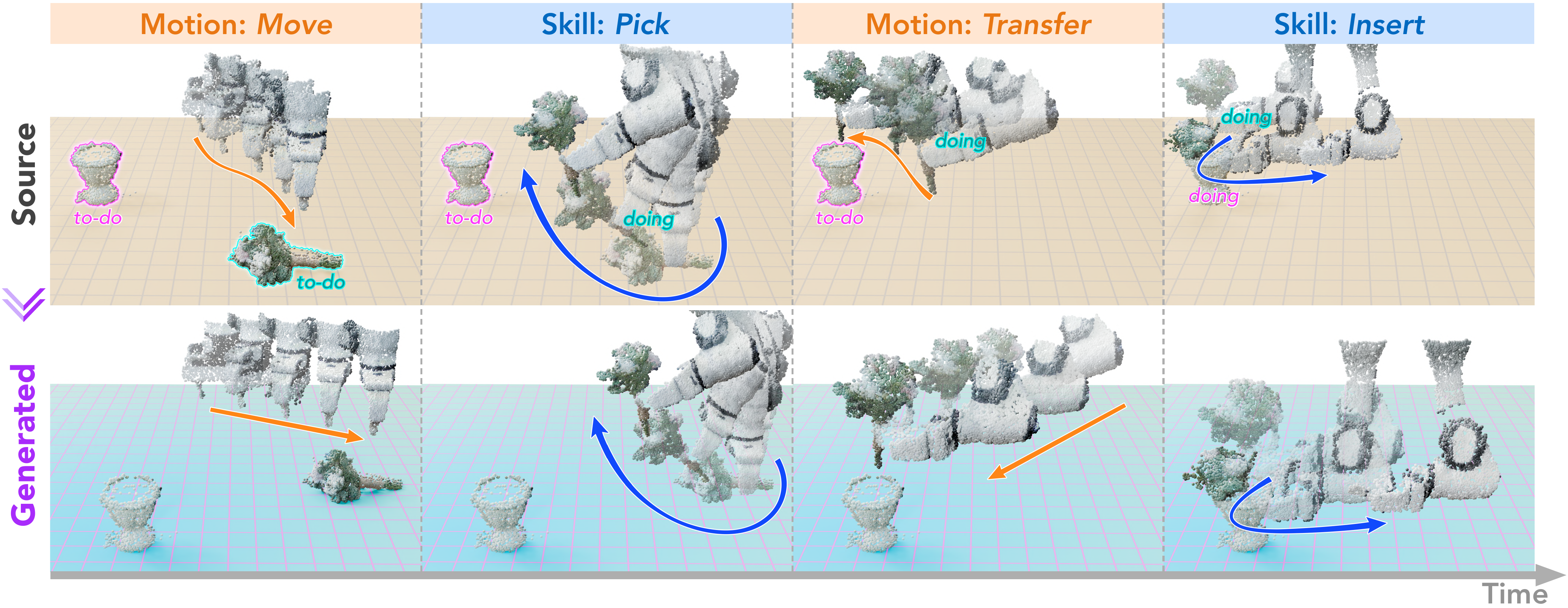}
    
    \caption{\textbf{Illustrations for synthetic visual observation generation.} Objects in the \textit{to-do} stage are segmented and transformed by the target object configurations. Objects in the \textit{doing} stage are merged with the end-effector and transformed according to the proprioceptive states.}
    \label{fig:method-traj}
    \vspace{-0.2cm}
\end{figure*}

\subsection{TAMP-based Action Generation}

\noindent\textbf{Adapting actions to the new configuration.} The generation process begins by selecting a target initial configuration $s_0'=\{\mathbf{T}_0^{{O_1}'}, \mathbf{T}_0^{{O_2}'}, ..., \mathbf{T}_0^{{O_K}'}\}$. Under the $4\times 4$ homogeneous matrix representation, the spatial transformation between the target and source configurations is computed as: 
\begin{align*}
    \Delta s_0
    &=\{(\mathbf{T}_0^{{O_1}})^{-1} \cdot \mathbf{T}_0^{{O_1}'}, \dots, (\mathbf{T}_0^{{O_K}})^{-1} \cdot \mathbf{T}_0^{{O_K}'}\}.
\end{align*}

Recall that the actions consist of both robot arm and robot hand commands. The robot hand commands define the interactive actions \textit{on} the object, e.g., holding the flower with the gripper, or rolling up the dough with the dexterous hand. Since they are \textit{invariant} of the spatial transformation, $a^{\mathrm{hand}}_t$ should remain unchanged regardless of the object configuration:
\begin{equation*}
    \hat{a}^{\mathrm{hand}}_t=a^{\mathrm{hand}}_t, \quad\forall~ t, s_o, s_0'.
\end{equation*}

In contrast, the robot arm commands should be spatially \textit{equivariant} to the object movements in order to adjust the trajectory according to the altered configuration.
Specifically, for the motion and skill segments involving the $k$-th object, we adapt the robot arm commands $\textcolor{myorange}{\mathbf{A}^\mathrm{EE}[{\bm{\tau}^{\mathrm{m}}_k}]}, \textcolor{myblue}{\mathbf{A}^\mathrm{EE}[{\bm{\tau}^{\mathrm{s}}_k}]}$ following a TAMP-based procedure, illustrated by Fig.~\ref{fig:method-motion-skill}.

For the skill segments with dexterous on-object behaviors, the spatial relations between end-effectors and objects must remain relatively static. Thus, the entire skill segments are transformed following the corresponding objects:
\begin{equation*}
    \textcolor{myblue}{\hat{\mathbf{A}}^\mathrm{EE}[{\bm{\tau}^{\mathrm{s}}_k}]}=\textcolor{myblue}{\mathbf{A}^\mathrm{EE}[{\bm{\tau}^{\mathrm{s}}_k}]} \cdot (\mathbf{T}_0^{{O_k}})^{-1} \cdot \mathbf{T}_0^{{O_k}'}.
\end{equation*}

For the motion segments moving in free space, the goal is to chain adjacent skill segments. Therefore, we plan the robot arm commands in the motion stage via motion planning:
\begin{equation*}
    \textcolor{myorange}{\hat{\mathbf{A}}^\mathrm{EE}[{\bm{\tau}^{\mathrm{m}}_k}]}=\texttt{MotionPlan}(\textcolor{myblue}{\hat{\mathbf{A}}^\mathrm{EE}[{\bm{\tau}^{\mathrm{s}}_{k-1}}][-1]},~ \textcolor{myblue}{\hat{\mathbf{A}}^\mathrm{EE}[{\bm{\tau}^{\mathrm{s}}_k}][0]}),
\end{equation*}
where the starting pose for motion planning is taken from the last frame of the previous skill segment, and the ending pose is from the first frame of the current skill segment. For simple uncluttered workspaces, linear interpolation suffices. For complex environments requiring obstacle avoidance, an off-the-shelf motion planning method~\cite{kuffner2000rrt} is employed.

\vspace{0.2cm} \noindent\textbf{Failure-free action execution.}
To ensure the validity of synthetic demonstrations without on-robot rollouts to filter out failed trajectories, we require failure-free action execution. Unlike previous works~\cite{mandlekar2023mimicgen, garrett2024skillmimicgen} that rely on operational space controllers and delta end-effector pose control, we employ inverse kinematics (IK) controllers~\cite{Zakka_Mink_Python_inverse_2024} and target absolute end-effector poses. Empirically, these adjustments are found to help minimize compounding control errors, contributing to the successful execution of the generated actions.

\subsection{Fully Synthetic Observation Generation}

\noindent\textbf{Adapting proprioceptive states.} 
The observations consist of point cloud data and proprioceptive states. Since the proprioceptive states share the same semantics with the actions, they should undergo the same transformation:
\begin{align*}
    \hat{o}^{\mathrm{hand}}_t&=o^{\mathrm{hand}}_t, \quad\forall~ t, s_o, s_0'; \\
    \hat{o}^{\mathrm{arm}}_t&=o^{\mathrm{arm}}_t \cdot ({\mathbf{A}}^\mathrm{EE}_t)^{-1} \cdot \hat{{\mathbf{A}}}^\mathrm{EE}_t.
\end{align*}
It is noteworthy that we found directly replacing the current state with the next target pose action (i.e., $\hat{o}^{\mathrm{arm}}_t \gets \hat{a}^{\mathrm{arm}}_{t+1}$) may impair performance, as the IK controllers may not always achieve the exact target pose.

\vspace{0.2cm} \noindent\textbf{Synthesizing point cloud observations.}
To synthesize the spatially augmented point clouds for the robot and objects, we employ a simple segment-and-transform strategy.
Apart from the target transformations, the only required information for synthesis is the segmentation masks for the $K$ objects on the first frame of the source demonstration, obtained in Sec.~\ref{sec:method-preprocess}.


For each object, we define $3$ stages. In the \textit{to-do} stage, the object is static and unaffected by the robot, and its point cloud is transformed according to the initial object configuration $(\mathbf{T}_0^{{O_k}})^{-1} \cdot \mathbf{T}_0^{{O_k}'}$. In the \textit{doing} stage, the object is in contact with the robot, and its point cloud is merged with the end-effector’s point cloud. In the \textit{done} stage, the object remains in its final state. These stages are easily identified by referencing the trajectory-level motion and skill segments.

For the robot’s end-effector, its point cloud undergoes the same transformation as indicated by the proprioceptive states
$({\mathbf{A}}^\mathrm{EE}_t)^{-1} \cdot \hat{{\mathbf{A}}}^\mathrm{EE}_t$. 
Given the assumption of a cropped workspace, the point clouds for the robot and the objects in the \textit{doing} stage can be separated by subtracting the object point clouds in the \textit{to-do} and \textit{done} stages from the scene point cloud.

A concrete example of this process is shown in Fig.~\ref{fig:method-traj}. More examples of the synthetic trajectories in real-world experiments can be found in Fig.~\ref{fig:traj-examples} in the appendix.

%% file: sections/5_sim.tex
\begin{figure*}
    \centering
    \includegraphics[width=1\linewidth]{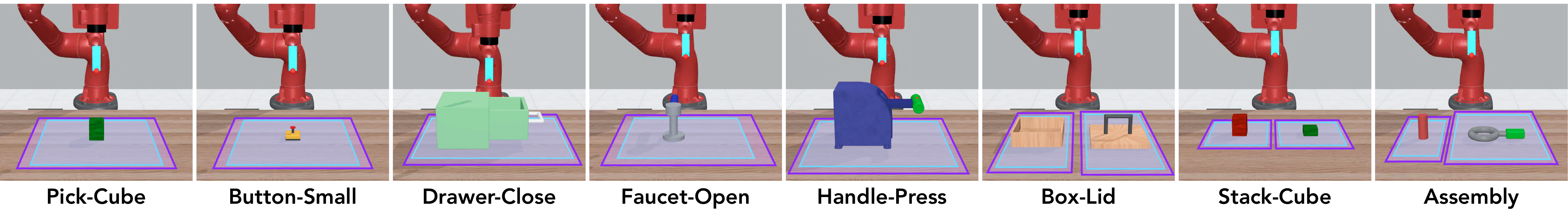}
    \vspace{-0.55cm}
    \caption{\textbf{Tasks for simulated evaluation on spatial generalization.} Purple and sky-blue rectangles mark the workspaces for demonstration generation and evaluation, respectively. The detailed sizes of these workspaces are listed in Tab.~\ref{table:sim-range} in the appendix.}
    \label{fig:sim-tasks}
\end{figure*}

\input{tabs/sim-results}

\section{Experiments in the Simulator}

\subsection{Effectiveness: One-Shot Imitation}
Before deploying \method to the real world, we evaluate its effectiveness in the simulator by training visuomotor policies on datasets generated by \method from only one source demonstration per task.

\vspace{0.2cm} \noindent\textbf{Policy.}
Both in the simulator and real world, we select DP3~\cite{ze20243d} as the visuomotor policy, which predicts actions by consuming point cloud and proprioception observations. For a fair comparison, we fix the total training steps counted by observation-action pairs for all evaluated settings, resulting in an equal training cost regardless of the dataset size.
The training details are listed in Appendix~\ref{sec:appendix-policy-training}.

\vspace{0.2cm} \noindent\textbf{Tasks.}
We design $8$ tasks adapted from the MetaWorld~\cite{yu2020metaworld} benchmark, illustrated in Fig.~\ref{fig:sim-tasks}. To strengthen the significance of spatial generalization, we modify these tasks to have enlarged object randomization ranges, as listed in Appendix~\ref{sec:appendix-sim-range}.

\vspace{0.2cm} \noindent\textbf{Generation and evaluation.}
We write scripted policies for these tasks and prepare only $1$ source demonstration per task for demonstration generation. We also produce $10$ and $25$ source demonstrations per task using the scripted policy as a reference for human-collected datasets. 
Based on the one source demonstration, we leverage \method to generate $100$ spatially augmented demonstrations for the tasks containing the spatial randomization of one object. Since the tasks concerning two objects have a more diverse range of object configurations, $200$ demonstrations are generated. 

\vspace{0.2cm} \noindent\textbf{Results analysis.}
The evaluation results for the simulated tasks are presented in Tab.~\ref{table:sim-result}. \method significantly enhances the policy performance compared with the source demonstration baseline. The policies trained on \method-generated datasets also outperform those trained on $10$ source demonstrations and get close to $25$ source demonstrations. This indicates \method has the potential to maintain the policy performance with over $20\times$ reduced human effort for data collection.

\subsection{Limitation: The Visual Mismatch Problem}
\label{sec:visual-mismatch}
\begin{figure}
    \vspace{-0.4cm}
    \centering
    \includegraphics[width=1\linewidth]{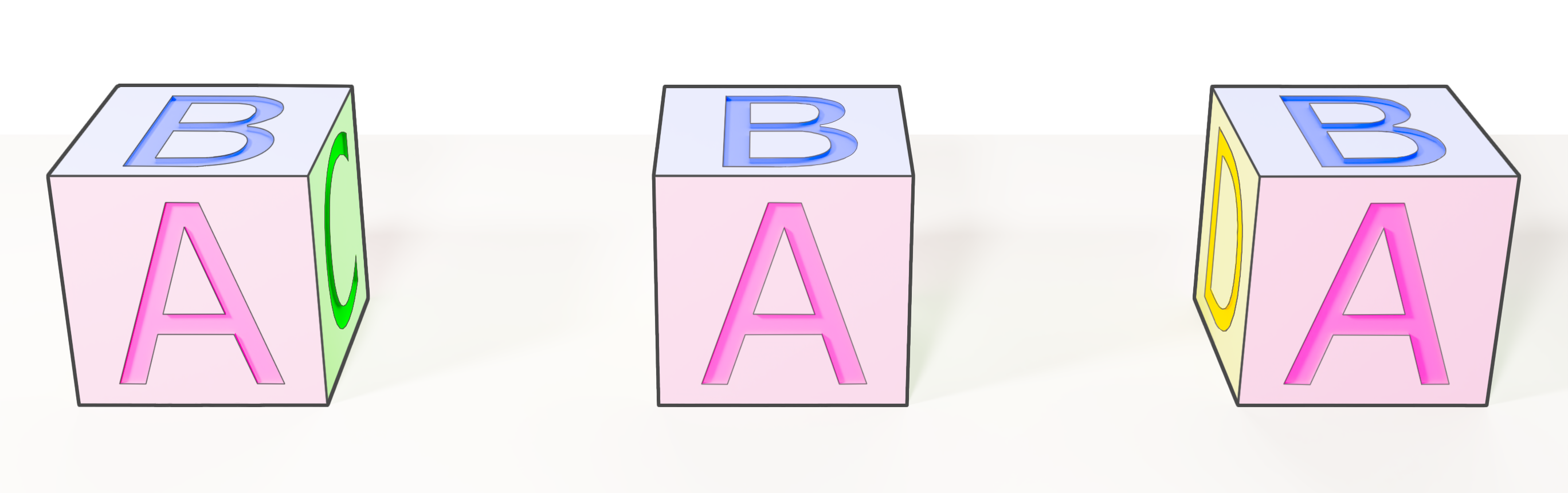}
    \vspace{-0.5cm}
    \caption{\textbf{Illustration for the visual mismatch problem.} As objects move through 3D space, their appearance changes due to variations in perspective. Under the constraint of a single-view observation, synthetic demonstrations consistently reflect a fixed side of the object's appearance seen in the source demonstration. This discrepancy causes a visual mismatch between the synthetic and real-captured data.}
    \label{fig:visual-mismatch}
\end{figure}

While the one-shot imitation experiment verifies the effectiveness of \method, it also reveals its limitation: synthetic demonstrations generated from one source demonstration are not as effective as the same number of human-collected demonstrations. We attribute the performance gap to the visual mismatch between the synthetic point clouds and those captured in the real world, under the constraint of a single-view observation perspective. An illustration is provided in Fig.~\ref{fig:visual-mismatch}. 

\begin{figure}
    \centering
    \vspace{-0.3cm}
    \includegraphics[width=1\linewidth]{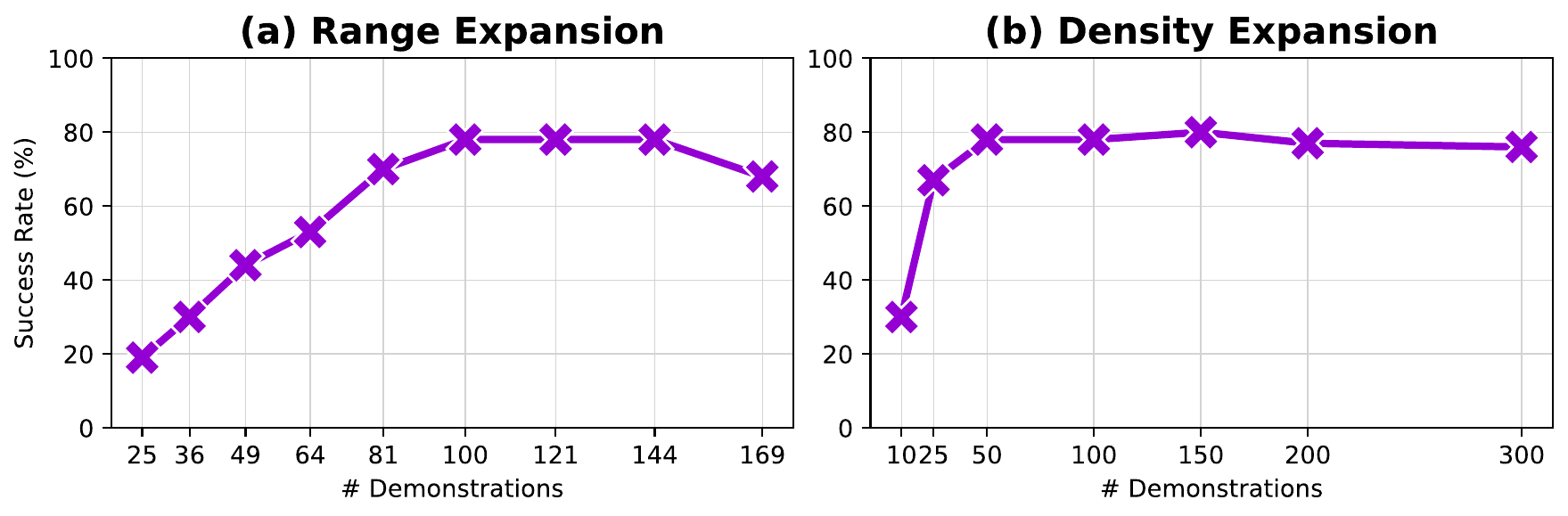}
    \caption{\textbf{Performance Saturation.} We report the policy performance boost w.r.t. the increase of synthetic demonstrations over $3$ seeds.}
    \label{fig:performance-saturation}
    \vspace{-0.3cm}
\end{figure}

\vspace{0.2cm} \noindent\textbf{Performance saturation.} A notable consequence of the visual mismatch problem is the phenomenon of performance saturation. 
An empirical analysis is conducted on the Pick-Cube task. 
In Fig.~\ref{fig:performance-saturation}(a), we fix the spatial density of target object configurations in the synthetic demonstrations and increase their spatial coverage by adding more synthetic demonstrations.
The curve indicates that the performance improvement plateaus once the spatial coverage exceeds a certain threshold. This saturation occurs because the visual mismatch intensifies as the distance between the source and synthetic object configurations increases, making additional synthetic demonstrations ineffective.
In Fig.~\ref{fig:performance-saturation}(b), a similar performance saturation effect is observed when we increase the density while keeping the spatial coverage fixed. This indicates excessive demonstrations are unnecessary once they sufficiently cover the workspace.

%% file: tabs/sim-results.tex
\begin{table*}[t]
\centering
\caption{\textbf{Simulated evaluation of \method for spatial generalization.} We report the maximum/averaged success rates over $3$ seeds.}
\vspace{-0.1cm}
\label{table:sim-result}

\resizebox{1.0\textwidth}{!}{%
\begin{tabular}{l|cccccccc|c}
\toprule



& Pick-Cube & Button-Small & Drawer-Close & Faucet-Open & Handle-Press & Box-Lid & Stack-Cube & Assembly & Averaged \\

\midrule

$1$ Source & $0/0$ & $4/4$ & $55/50$ & $39/23$ & $17/16$ & $11/11$ &  $0/0$ & $0/0$ & $16/13$ \\

\textbf{\method} &  \ccbf{76/73} & \ccbf{92/84} & \ccbf{100/100} & \ccbf{95/92} & \ccbf{100/100} & \ccbf{100/95} & \ccbf{79/77} & \ccbf{86/83} & \ccbf{91/88}\\

\midrule

\textcolor{gray}{$10$ Source} & \gray{29/29} & \gray{54/52} & \gray{100/100} & \gray{90/89} & \gray{100/99} & \gray{94/89} &  \gray{44/38} & \gray{47/45} & \gray{70/68} \\

\textcolor{gray}{$25$ Source} & \gray{82/74} & \gray{90/84} & \gray{100/100} & \gray{100/100} & \gray{100/100} & \gray{100/100} &  \gray{95/93} & \gray{83/79} & \gray{94/91} \\

\bottomrule
\end{tabular}}
\vspace{-0.2cm}
\end{table*}

%% file: sections/6_real.tex
\begin{figure*}
    \centering
    \includegraphics[width=1\linewidth]{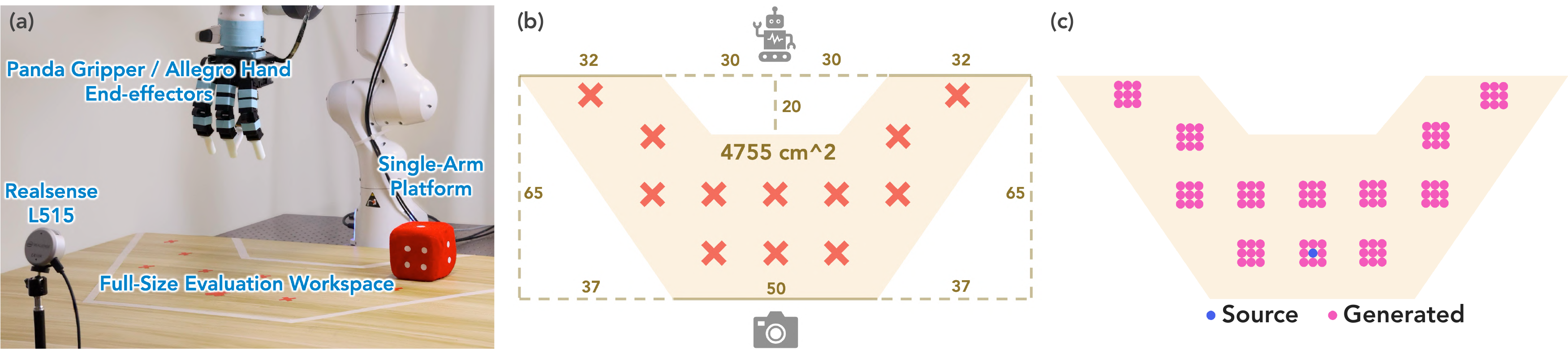}
    \caption{\textbf{Protocol for evaluating spatial generalization.} (a) Setups on the single-arm platform. (b) Illustration for the full-size evaluation workspace. (c) Illustration for the generation strategy targeting the evaluated configurations along with small-range perturbations.}
    \label{fig:real-spatial-setup}
\end{figure*}

\begin{figure*}
    \centering
    \includegraphics[width=1\linewidth]{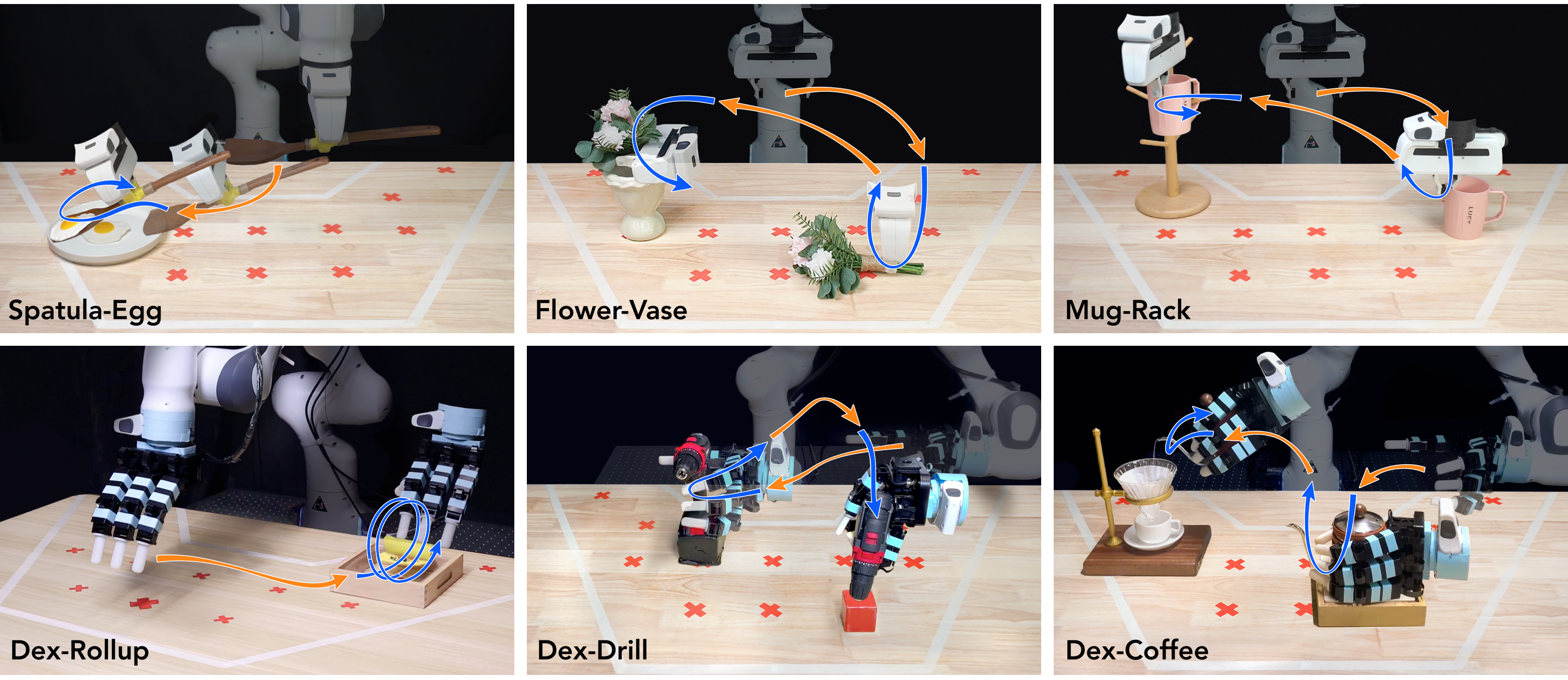}
    \caption{\textbf{Tasks for real-world evaluation on spatial generalization.} Spatula-Egg and Dex-Rollup are one-stage tasks involving contact-rich behaviors. Flower-Vase, Mug-Rack, Dex-Drill, and Dex-Coffee are two-stage tasks requiring precise manipulation.}
    \label{fig:real-spatial-tasks}
\end{figure*}

\section{Real-World Experiments: Spatial Generalization}
We assess the spatial generalization capability of visuomotor policies enhanced by \method across $8$ real-world tasks deployed on $3$ different platforms. $7$ tasks are performed on single-arm platforms with parallel grippers or dexterous hand end-effectors. Additionally, one task is executed on a bimanual humanoid. A task summary is provided in Tab.~\ref{table:task-summary}.

\input{tabs/task_summary}

\subsection{Single-Arm Platforms}

\noindent\textbf{Tasks.}
On the Franka Panda single-arm platform, we design $3$ tasks using the original Panda gripper and $4$ tasks using an Allegro dexterous hand as the end-effector. The \textcolor{myorange}{motion} and \textcolor{myblue}{skill} trajectories of these tasks are visualized in Fig.~\ref{fig:real-spatial-tasks} and the task descriptions are provided in Appendix~\ref{sec:appendix-task-real}.
For all tasks, a single Intel Realsense L515 camera is adopted to capture point cloud observations, as depicted in Fig.~\ref{fig:real-spatial-setup}(a).

\vspace{0.2cm} \noindent\textbf{Evaluation protocol.} 
To evaluate spatial generalization, we define a large planar evaluation workspace, the size of which corresponds to the maximum reach of the robot arm. We uniformly sample $12$ points within this irregularly-shaped workspace as the coordinates for potential object configurations, with a $15\textrm{cm}$ spacing between neighboring coordinates, as illustrated in Fig.~\ref{fig:real-spatial-setup}(b).

To determine the actual evaluated configurations for each task, we perform manual trials using kinematic teaching to confirm the feasibility of each configuration. For example, in the Dex-Rollup task, the dexterous hand can reach a piece of plasticine placed in the near-robot corner of the workspace with a vertical wrist angle. However, it cannot grasp a kettle in the same location using a horizontal wrist angle, as required in the Dex-Coffee task. We conduct trials on all feasible configurations and repeat the evaluations $5$ times per configuration to ensure the reliability of the results.

\input{tabs/real-spatial-results}

\begin{figure*}
    \vspace{-0.3cm}
    \centering
    \includegraphics[width=1\linewidth]{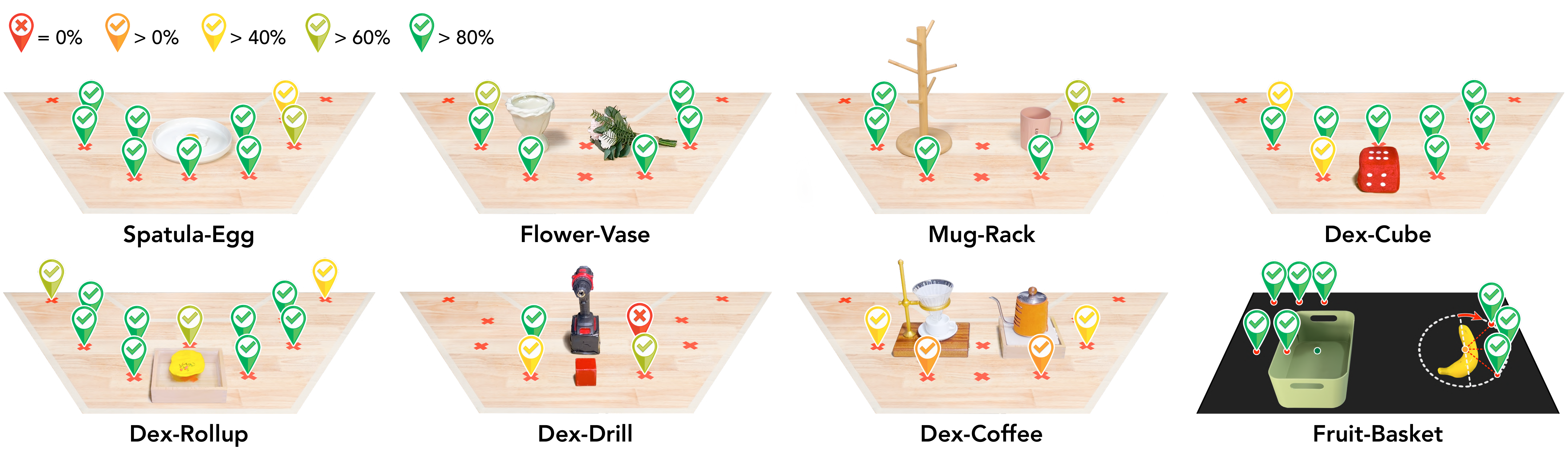}
    \caption{\textbf{Spatial heatmaps for the real-world evaluation results.} 
    The success rate for each coordinate is calculated as the average across all relevant trials. For example, each coordinate of the vase in the Flower-Vase task is in combination with $4$ coordinates of the flower, including the one appearing in the source demonstration. This results in a total of $20$ trials, given $5$ repetitions per combination.}
    \label{fig:real-success-heatmap}
    \vspace{-0.2cm}
\end{figure*}

\vspace{0.2cm} \noindent\textbf{Generation strategy.}
As in the simulated environments, we collect only one source demonstration for each task. However, real-world point cloud observations are often noisy, with issues such as flickering holes in the point clouds or projective smearing around object outlines. Even after performing clustering and downsampling during the point cloud preprocessing stage (Sec.~\ref{sec:method-preprocess}), the imitation learning policy can overfit to these irregularities if only one demonstration is provided.

To mitigate this issue, we replay the source demonstration twice and capture the corresponding point cloud observations. The altogether $3$ point cloud trajectories enrich the diversity in visual degradations and help alleviate the overfitting problem. 
Since replaying twice is low-cost, we consider this approach a beneficial tradeoff between efficiency and effectiveness.

For each task, we set the generated object configurations to correspond to the evaluated configurations.
However, human operators cannot always place objects with perfect precision in the real world, yet we found visuomotor policies are sensitive to even small deviations. Thus, we further augment the generated object configurations by adding small-range perturbations. Specifically, for each target configuration, we generate $9$ demonstrations with $(\pm 1.5 \textrm{cm}) \times (\pm1.5 \textrm{cm})$ perturbation to mimic slight placement variations in the real world. The final generated configurations are shown in Fig.~\ref{fig:real-spatial-setup}(c).

In summary, the total number of generated demonstrations is calculated as $3 \times (\mathrm{\# Eval}) \times 9$, which represents the $3$ source demonstrations, multiplied by the number of evaluated configurations, and further multiplied by the $9$ perturbations. The detailed counts for each task are listed in Tab.~\ref{table:task-summary}.

\vspace{0.2cm} \noindent\textbf{Results analysis.}
The performance of visuomotor policies~\cite{ze20243d} trained on $3$ source demonstrations and \method-generated demonstrations are reported in Tab.~\ref{table:real-spatial-result}. Agents trained solely on source demonstrations exhibit severe overfitting behaviors, blindly replicating the demonstrated trajectory. 
In Appendix~\ref{sec:appendix-increase-source}, we evaluate the policy performance trained on datasets containing additional human-collected demonstrations. We found the spatial effective range of the trained policies is upper-bounded by the sum of demonstrated configurations, aligned with the findings in the empirical study in Sec.~\ref{sec:empirical}.

Similar to the effects of manually covering the workspace with human-collected demonstrations, \method-generated datasets enable the agents to display a more adaptive response to diverse evaluated configurations, resulting in significantly higher success rates. \method consistently enhances the performance across all the evaluated tasks. Although the performance gains are less pronounced in the Dex-Drill and Dex-Coffee tasks, we found the policies trained on the generated data still guide the dexterous hands to generally appropriate manipulation poses. The relatively lower performance is primarily due to stringent precision requirements.


To further investigate the generalization capabilities enabled by \method, we visualize the spatial heatmaps for the evaluated configurations in Fig.~\ref{fig:real-success-heatmap}. The heatmaps reveal high success rates on configurations close to the demonstrated ones, while the performance diminishes as the distance from the demonstrated configuration increases. We attribute this decline to the visual mismatch problem caused by single-view observations, as previously discussed in Sec.~\ref{sec:visual-mismatch}.

A notable observation arises in the Dex-Rollup task, where the policy trained on the \method-generated dataset could dynamically adjust the number of wrapping motions ranging from $2$ to $5$ in response to the distinct plasticity of every hand-molded piece of plasticine. This suggests the usage of \method is not in conflict with the resulting agent's closed-loop re-planning capability. The intrinsic strength of visuomotor policies is effectively preserved.


\input{tabs/generation_cost}

\vspace{0.2cm} \noindent\textbf{Generation cost.}
We compare the time cost of real-world demonstration generation between MimicGen~\cite{mandlekar2023mimicgen} and \method. We estimate MimicGen's time cost by multiplying the duration of replaying a source trajectory by the number of generated demonstrations and adding an additional $20$ seconds per trajectory for human operators to reset the object configurations. It is important to note that MimicGen involves continuous human intervention, while the time cost of \method is purely computational, without the involvement of either the robot or human operators.


\subsection{Bimanual Humanoid Platform}

\begin{figure}
    \centering
    \includegraphics[width=0.99\linewidth]{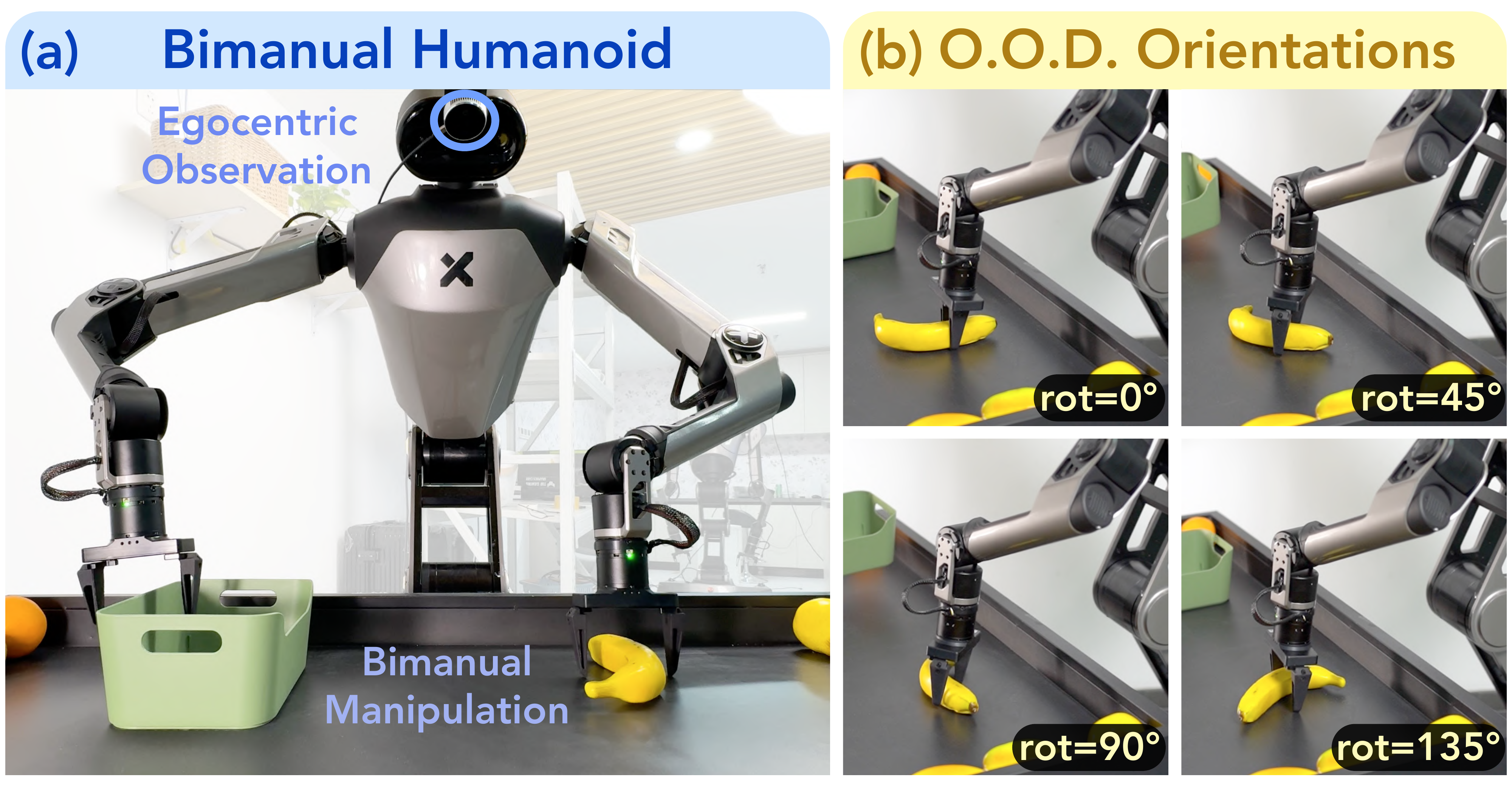}
    \caption{\textbf{Bimanual humanoid platform.} (a) Egocentric observations and bimanual manipulation. (b) The Fruit-Basket task involves the out-of-distribution orientations during evaluation.}
    \label{fig:real-humanoid}
    \vspace{-0.3cm}
\end{figure}

\noindent\textbf{Task.}
In addition to the tasks on the single-arm platform, we also designed a Fruit-Basket task on a Galaxea R1 robot, illustrated in Fig.~\ref{fig:real-humanoid}. The Fruit-Basket task is distinguished from the previous tasks by three key features: 

\textit{1) Bimanual manipulation.} The robot simultaneously grasps the basket with one arm and the banana with the other. The right arm then places the basket in the center of the workspace, while the left arm places the banana into the basket. 

\textit{2) Egocentric observation.} The camera is mounted on the robot's head~\cite{ze2024generalizable}. While the robot’s base is immobilized in this task, the first-person view opens opportunities for future deployment in mobile manipulation scenarios. 

\textit{3) Out-of-distribution orientations.} Still using a single human-collected demonstration, the banana is placed with orientational offsets (i.e., $45^\circ$, $90^\circ$, and $135^\circ$) relative to the original demonstration during evaluation, while the basket position is randomized within a $10\,\textrm{cm} \times 5\,\textrm{cm}$ workspace.

\vspace{0.2cm} \noindent\textbf{Generation strategy.}
The generation procedure follows a similar approach as that used for the single-arm platform. Specifically, the human-collected demonstration is replayed twice, yielding $3$ source demonstrations in total. \method generates synthetic demonstrations by independently adapting the actions of both arms to the respective transformations of the objects. Small-range perturbations are omitted in this task due to the relatively lower precision requirements.

A challenge in synthesizing point cloud observations with orientational offsets lies in the limited view provided by the single camera, which only captures the objects' front-facing appearance. To address this limitation, the humanoid robot adopts a stooping posture, enabling a near bird’s-eye view perspective. This adjustment allows for more effective point cloud editing to simulate full-directional yaw rotations.

\vspace{0.2cm} \noindent\textbf{Results analysis.} 
The success rates for both the source and generated datasets are compared in Tab.~\ref{table:real-spatial-result}, and the spatial heatmap is shown in Fig.~\ref{fig:real-success-heatmap}. The high success rate of $90.8\%$ demonstrates the effectiveness of \method on bimanual humanoid platforms and its ability to help policies generalize to out-of-distribution orientations. 
A more detailed analysis is presented in Appendix~\ref{sec:appendix-humanoid}.

\begin{figure}[bp]
    \centering
    \vspace{-0.2cm}
    \includegraphics[width=1\linewidth]{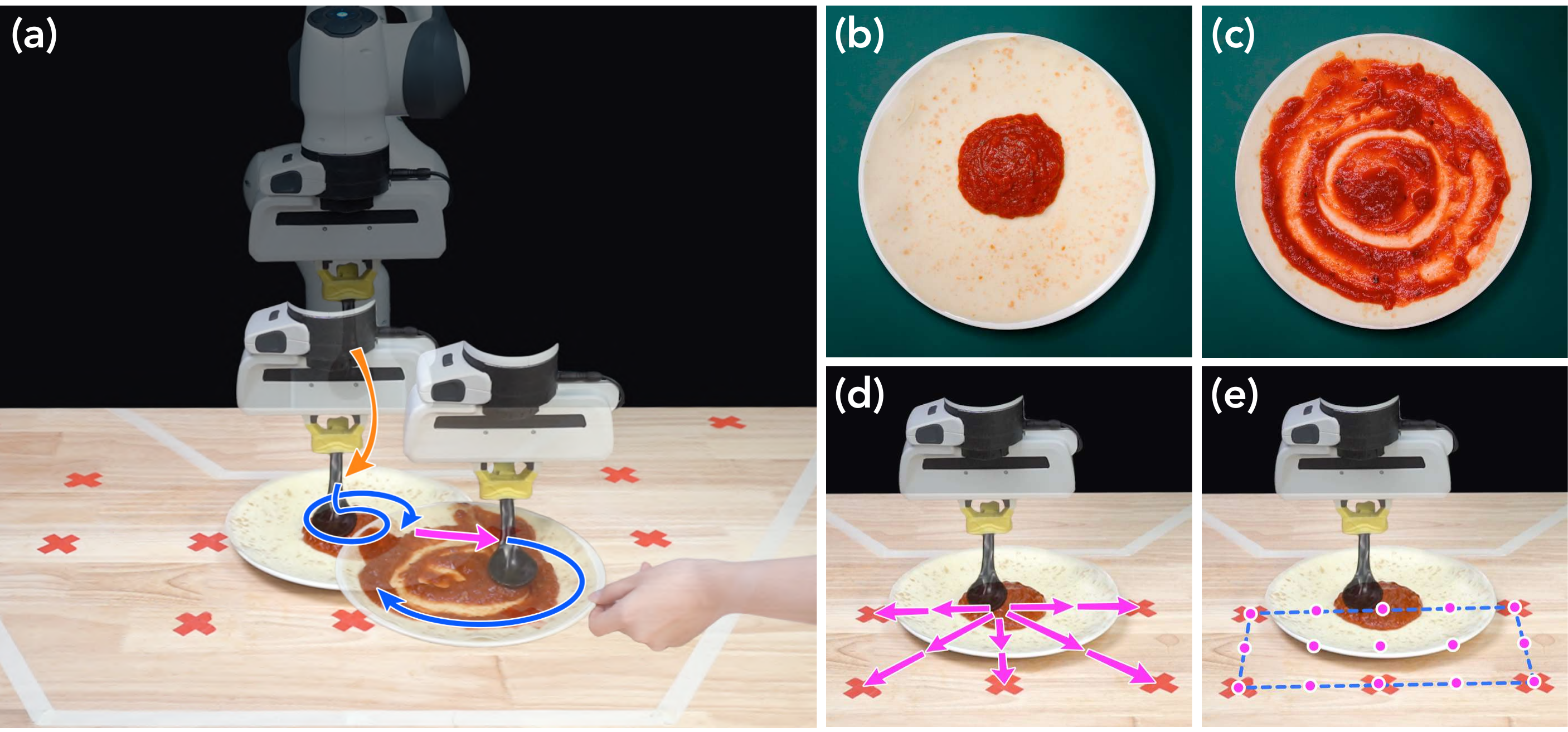}
    \caption{\textbf{\method for disturbance resistance.} (a-c) Illustration, initial, and ending states of the Sauce-Spreading task. (d) Disturbance applied for quantitative evaluation. (e) Standard generation strategy. }
    \label{fig:real-sauce-setup}
\end{figure}

\section{Real-World Experiments: Extensions}

\subsection{Disturbance Resistance}


One critical advantage of visuomotor policies is their ability to perform closed-loop corrections under disturbances. We investigate whether a \method-generated dataset, derived from one human-collected and two replayed source demonstrations, can train visuomotor policies equipped with such capability.

\vspace{0.2cm} \noindent\textbf{Task.}
We consider a Sauce-Spreading task (Fig.~\ref{fig:real-sauce-setup}(a)) adapted from DP~\cite{chi2023diffusion_policy}. 
Initially, the pizza crust contains a small amount of sauce at its center (Fig.~\ref{fig:real-sauce-setup}(b)). 
The gripper maneuvers the spoon in hand to approach the sauce center and periodically spread it to cover the pizza crust in a spiral pattern (Fig.~\ref{fig:real-sauce-setup}(c)).


\vspace{0.2cm} \noindent\textbf{Evaluation protocol.} During the sauce-spreading process, disturbances are introduced by shifting the pizza crust twice to the neighboring spots within the workspace. We consider $5$ neighboring spots (Fig.~\ref{fig:real-sauce-setup}(d)) and conduct $5$ trials per spot, resulting in $25$ trials.
For quantitative evaluation, we measure the sauce coverage on the pizza crust. Additionally, we report a normalized sauce coverage score, where $0$ represents no operation taken, and $100$ corresponds to human expert performance. Detailed calculations are provided in Appendix~\ref{sec:appendix-disturb}.

\begin{figure}
    \centering
    \includegraphics[width=0.8\linewidth]{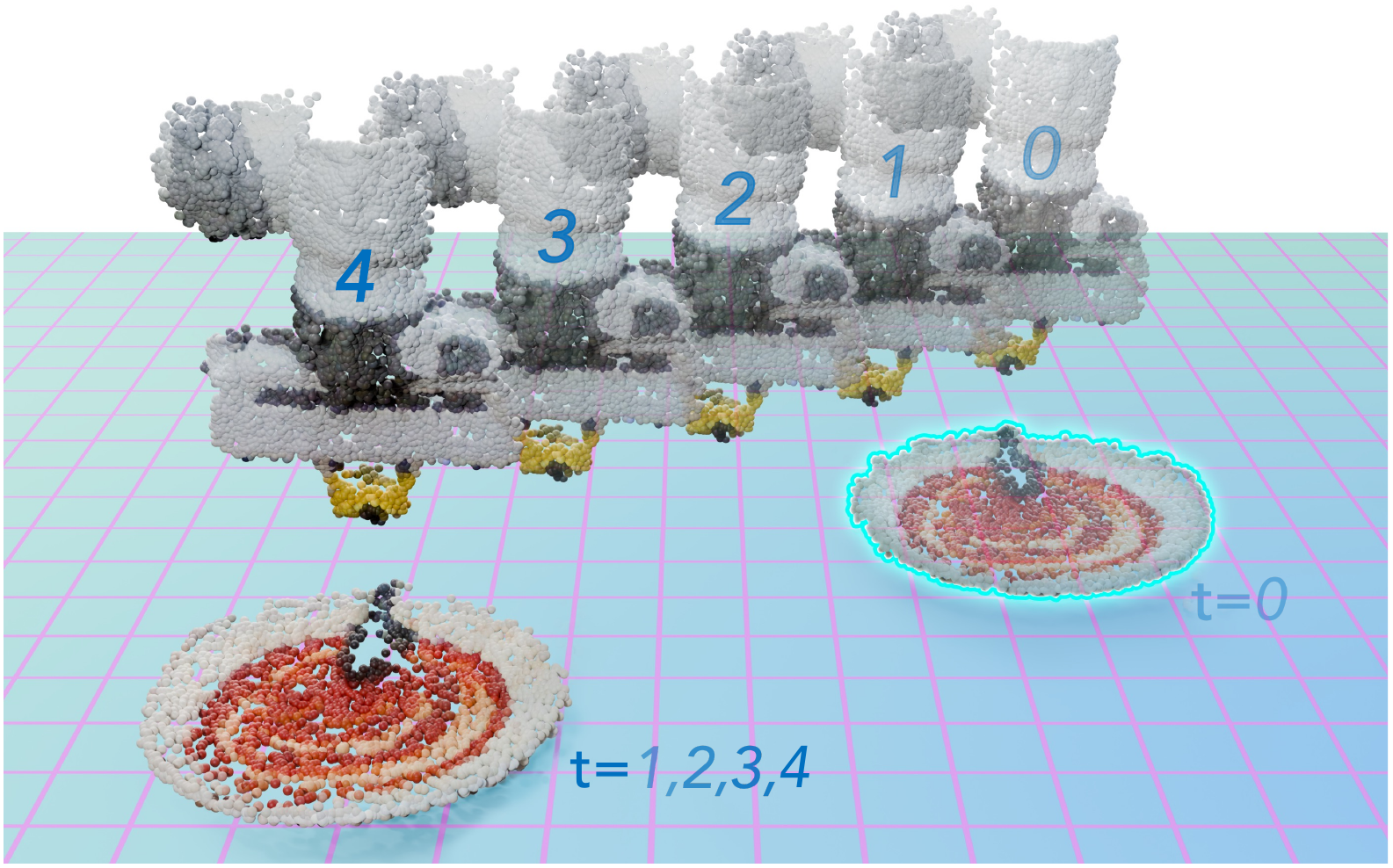}
    \caption{\textbf{Illustration for the ADR strategy. } Asynchronous transformations are applied to the disturbed object and the robot end-effector to simulate the disturbance resistance process.}
    \label{fig:real-sauce-gen}
\end{figure}



\vspace{0.2cm} \noindent\textbf{Generation strategies.} A standard generation strategy selects $15$ intermediate spots (Fig.~\ref{fig:real-sauce-setup}(e)) observed during the disturbance process as the initial object configurations for a standard \method data generation procedure.

To specifically enhance disturbance resistance, we propose a specialized strategy named Augmentation for Disturbance Resistance (ADR), illustrated in Fig.~\ref{fig:real-sauce-gen}. In ADR, the pizza crust is artificially displaced to nearby positions at certain time steps to simulate the disturbance. The robot's end-effector, holding the spoon, initially remains static and subsequently interpolates its motion to re-approach the displaced crust before continuing the periodic spreading motion.

\input{tabs/real-disturb}


\vspace{0.2cm} \noindent\textbf{Results analysis.} Tab.~\ref{table:real-disturb} presents the sauce coverage and normalized scores for both the standard \method and the ADR-enhanced \method strategies. The ADR strategy significantly outperforms the standard \method, achieving performance comparable to human experts.
In the video, we showcase the ADR-enhanced policy is still robust under up to $5$ successive disturbances.
These findings underscore the critical role of the demonstration data in enabling policy capabilities. The ability to resist disturbances does not emerge naturally but is acquired through targeted disturbance-involved demonstrations.




\subsection{Obstacle Avoidance}
\begin{figure}
    \centering
    \includegraphics[width=\linewidth]{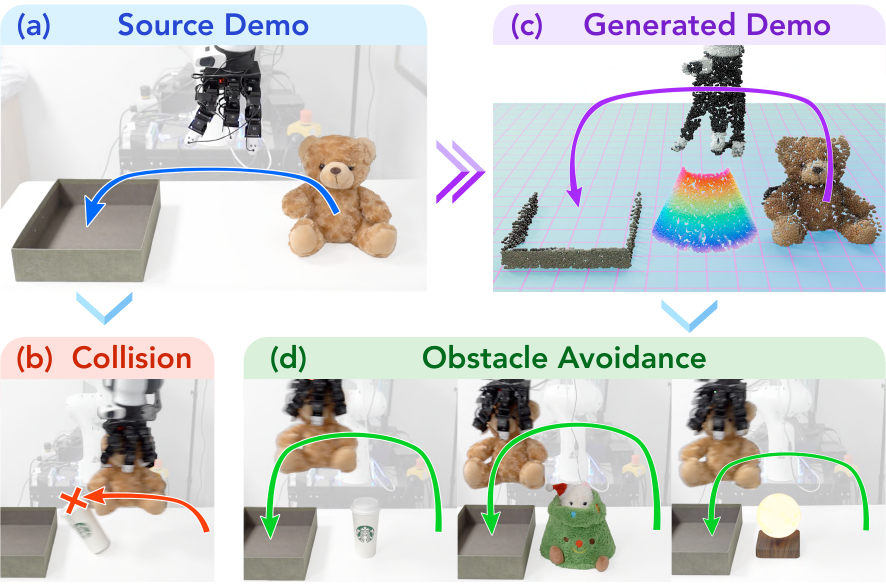}
    \caption{\textbf{\method for obstacle avoidance.} (ab) Policy trained on the source demonstration collides with the unseen obstacle. (cd) Policy trained on the generated dataset could avoid diverse-shaped obstacles.}
    \label{fig:real-obstacle}
    \vspace{-0.2cm}
\end{figure}

\noindent\textbf{Task.}
The ability to avoid obstacles is also imparted through demonstrations containing obstacle-avoidance behaviors. To investigate such capability, we introduce obstacles to a Teddy-Box task, where the dexterous hand grasps the teddy bear and transfers it into the box on the left (Fig.~\ref{fig:real-obstacle}(a)). Trained on the source demonstrations without obstacles, the visuomotor policy fails to account for potential collisions, e.g., it might knock over the coffee cup placed in the middle (Fig.~\ref{fig:real-obstacle}(b)). 

\vspace{0.2cm} \noindent\textbf{Generation strategy.}
To generate obstacle-involved demonstrations, we augment the real-world point cloud observations by sampling points from simple geometries, such as boxes and cones, and fusing these points into the original scene (Fig.~\ref{fig:real-obstacle}(c)). Obstacle-avoiding trajectories are generated by a motion planning tool~\cite{kuffner2000rrt}, ensuring collision-free actions.

\vspace{0.2cm} \noindent\textbf{Evaluation and results analysis.}
For evaluation, we position $5$ everyday objects with diverse shapes in the middle of the workspace (Fig.~\ref{fig:real-obstacle}(d)) and conduct $5$ trials per object, resulting in a total of $25$ trials. The agent trained on the augmented dataset successfully bypasses obstacles in $22$ out of $25$ trials. Notably, in scenarios without obstacles, the agent follows the lower trajectory observed in the source demonstrations, indicating its responsiveness to environmental variations.

%% file: tabs/task_summary.tex
\begin{table}[t]
\caption{\textbf{A summary of real-world tasks for spatial generalization evaluation.}
ActD: action dimension. \#Obj: number of manipulated objects. \#Eval: number of evaluated configurations. \#GDemo: number of \method-generated demonstrations.}
\label{table:task-summary}
\vspace{-0.05in}

\resizebox{0.5\textwidth}{!}{%
\begin{tabular}{lccccccccc}
\midrule

Task & Platform &  ActD & \#Obj & \#Eval & \#GDemo \\

\midrule
\textbf{Spatula-Egg} & Gripper & $6$ & $1$ & $10$ & $270$ \\

\textbf{Flower-Vase} & Gripper & $7$ & $2$ & $4 \!\times\! 4$ & $432$ \\

\textbf{Mug-Rack} & Gripper & $7$ & $2$ & $4 \!\times\! 4$ & $432$ \\

\textbf{Dex-Cube} & Dex. Hand & $22$ & $1$ & $10$ & $270$ \\

\textbf{Dex-Rollup} & Dex. Hand & $22$ & $1$ & $12$ & $324$\\

\textbf{Dex-Drill} & Dex. Hand & $22$ & $2$ & $3 \!\times\! 3$ & $243$ \\

\textbf{Dex-Coffee} & Dex. Hand & $22$ & $2$ & $3 \!\times\! 3$ & $243$ \\

\textbf{Fruit-Basket} & Bimanual & $14$ & $2$ & $4 \!\times\! 6$ & $72$ \\


\bottomrule
\end{tabular}
}
 \end{table}

%% file: tabs/real-spatial-results.tex
\begin{table*}[t]
\centering
\caption{\textbf{Real-world evaluation of \method for spatial generalization.} For reliable evaluation, a total of $530$ policy rollouts are conducted on the $8$ tasks. The success rate for each task is averaged on $5$ repetitions for each evaluated configuration. The evaluated configurations for each task are counted in Tab.~\ref{table:task-summary}, and visualized in Fig.~\ref{fig:real-success-heatmap}.}
\label{table:real-spatial-result}
\resizebox{1.0\textwidth}{!}{%
\begin{tabular}{l|cccccccc|c}
\toprule



 & Spatula-Egg & Flower-Vase & Mug-Rack & Dex-Cube & Dex-Rollup & Dex-Drill & Dex-Coffee & Fruit-Basket & Averaged \\

\midrule

Source & $10.0$ & $6.3$ & $6.3$ & $10.0$ & $8.3$ &  $11.1$ & $11.1$ & $25.0$ & $11.0$ \\

\textbf{\method} &  \ccbf{88.0} & \ccbf{82.5} & \ccbf{85.0} & \ccbf{78.0} & \ccbf{76.7} & \ccbf{55.6} & \ccbf{40.0} & \ccbf{90.8} & \ccbf{74.6}\\

\bottomrule
\end{tabular}}
\end{table*}

%% file: tabs/generation_cost.tex
\begin{table}[h]
\centering
\caption{\textbf{The time cost for generating real-world demonstrations.} The computational cost of \method is measured on a single-process procedure. Since the synthetic generation process is highly parallelizable, it can be further accelerated using multi-processing.}
\label{table:real-disturb}
\resizebox{1.0\linewidth}{!}{%
\begin{tabular}{l|ccc}
\toprule

  & Single o-a Pair & A Trajectory & Whole Dataset \\

\midrule

MimicGen & \cc{2.1\,\mathrm{s}} & \cc{2.1\,\mathrm{min}} & \cc{83.7\,\mathrm{h}} \\

\textbf{\method} & \ccbf{0.00015\,\mathrm{s}} & \ccbf{0.010\,\mathrm{s}} & \ccbf{22.0\,\mathrm{s}} \\

\bottomrule
\end{tabular}}
\vspace{-0.3cm}
\end{table}

%% file: tabs/real-disturb.tex
\begin{table}[t]
\centering
\vspace{-0.1cm}
\caption{\textbf{Real-world evaluation of \method for disturbance resistance.} Raw evaluation results and detailed definitions for the metrics are presented in Appendix~\ref{sec:appendix-disturb}.}
\label{table:real-disturb}
\resizebox{\linewidth}{!}{%
\begin{tabular}{l|cc}
\toprule

  & Sauce Coverage & Normalized Score \\

\midrule

Regular \method & \cc{34.2} & \cc{40.4} \\

\textbf{\method w/ ADR \quad} & \ccbf{61.2} & \ccbf{92.3} \\

\midrule

\textcolor{gray}{Initial State} & \gray{13.2} & \gray{0} \\
\textcolor{gray}{Human Expert} & \gray{65.2} & \gray{100}\\

\bottomrule
\end{tabular}}
\vspace{-0.5cm}
\end{table}

%% file: sections/9_conclusion.tex
\section{Conclusion}

In this work, we introduced \method, a fully synthetic data generation system designed to facilitate visuomotor policy learning by mitigating the need for large volumes of human-collected demonstrations. 
Through TAMP-based action adaption and 3D point cloud manipulation, \method enables the generation of spatially augmented demonstrations with minimal computational cost, significantly improving spatial generalization and policy performance across a wide range of real-world tasks and platforms.
Furthermore, we extend \method to generate demonstrations incorporating disturbance resistance and obstacle avoidance behaviors, endowing the trained policies with the corresponding capabilities. 


\vspace{0.2cm} \noindent\textbf{Limitations.} 
Although we have demonstrated the effectiveness of \method, it has several limitations.
First, \method relies on the availability of segmented point clouds, which limits its applicability in highly cluttered or unstructured environments.
Second, \method is not suitable for tasks where spatial generalization is not required, such as in-hand reorientation~\cite{chen2022system} or push-T~\cite{florence2022ibc,chi2023diffusion_policy} with a fixed target pose.
Third, the performance of \method is affected by the visual mismatch problem, as previously discussed in Sec.~\ref{sec:visual-mismatch}. 

\vspace{0.2cm} \noindent\textbf{Future works.} 
Future works could explore mitigating the impact of visual mismatch, potentially by leveraging techniques such as contrastive learning or 3D generative models.
Another avenue for future research is to use additional human-collected demonstrations as source data, aiming to identify the optimal balance between policy performance and the overall cost of data collection.

\section*{Acknowledgement}
We would like to give special thanks to Galaxea Inc. for providing the R1 robot and Jianning Cui, Ke Dong, Haoyin Fan, and Yixiu Li for their technical support. We also thank Gu Zhang, Han Zhang, and Songbo Hu for hardware setup and data collection, Yifeng Zhu and Tianming Wei for discussing the controllers in the simulator, and Widyadewi Soedarmadji for the presentation advice.
Tsinghua University Dushi Program supports this project.

%% file: sections/X_appendix.tex
\clearpage
\newpage
\begin{appendix}

\subsection{Policy Training and Implementation Details}
\label{sec:appendix-policy}

We select 3D Diffusion Policy (DP3)~\cite{ze20243d} as the visuomotor policy used for real-world and simulated experiments. We compare its performance against 2D Diffusion Policy (DP)~\cite{chi2023diffusion_policy} in the empirical study in Sec.~\ref{sec:empirical}. We list the training and implementation details as follows.

\vspace{0.2cm}\subsubsection{Details for Policy Training} 
\label{sec:appendix-policy-training}
For a fair comparison, we fix the total training steps counted by observation-action pairs to be $2\mathrm{M}$ for all evaluated settings, resulting in an equal training cost regardless of the dataset size. 
To stabilize the training process, we use AdamW~\cite{loshchilov2017decoupled} optimizer and set the learning rate to be $1\mathrm{e}^{-4}$ with a $500$ step warmup.

In real-world experiments, we use the DBSCAN~\cite{ester1996density} clustering algorithm to discard the outlier points and downsample the number of points in the point cloud observations to $1024$. In the simulator, we skip the clustering stage and downsample the point clouds to $512$ points.

We follow the notation in the Diffusion Policy~\cite{chi2023diffusion_policy} paper, where $T_\mathrm{o}$ denotes the observation horizon, $T_\mathrm{p}$ as the action prediction horizon, and $T_\mathrm{a}$ denotes the action execution horizon. 
In real-world experiments, we set $T_\mathrm{o}=2,\,T_\mathrm{p}=8,\,T_\mathrm{a}=5$. We run the visuomotor policy at $10\mathrm{Hz}$. Since $T_\mathrm{a}$ indicates the steps of actions executed on the robot without re-planning, our horizon settings result in a closed-loop re-planning latency of $0.5$ seconds, responsive enough for conducting dexterous retrying behaviors and disturbance resistance.
In the simulator, since the tasks are simpler, we set $T_\mathrm{o}=2,\,T_\mathrm{p}=4,\,T_\mathrm{a}=3$.

\vspace{0.2cm}\subsubsection{Pre-Trained Encoders for Diffusion Policies} 
\label{sec:appendix-policy-pretrain}
To replace the train-from-scratch ResNet18~\cite{he2016deep} visual encoder in the original Diffusion Policy architecture, we consider $3$ representative pre-trained encoders: R3M~\cite{nair2023r3m}, DINOv2~\cite{oquab2023dinov2}, and CLIP~\cite{radford2021learning}.
R3M utilizes a ResNet~\cite{he2016deep} architecture and is pre-trained on robotics-specific tasks. DINOv2 and CLIP employ ViT~\cite{dosovitskiy2021an} architectures and are pre-trained on open-world vision tasks. These encoders are widely used in previous works~\cite{chi2024umi,lin2024data} to enhance policy performance.

\subsection{Spatial Generalization Empirical Study Details}
\label{sec:appendix-empirical}

In Sec.~\ref{sec:empirical}, we conducted an empirical study on the spatial generalization capability of visuomotor policies. In this section, we provide more detailed analysis of the study's results.

\vspace{0.2cm}\subsubsection{More Analysis on the Visualization Results in Fig.~\ref{fig:spatial_gen_vis}}
\label{sec:appendix-empirical-visualize}
The results suggest that visuomotor policies exhibit some degree of spatial interpolation capability. Specifically, the green-colored effective range in the \texttt{sparse} setting with $9$ demonstrations is significantly larger than $9$ times the effective range in the \texttt{single} setting. However, increased precision requirements would make it harder to interpolate, as indicated by comparing the two task variants under the \texttt{sparse} setting.

Meanwhile, extrapolation proves to be more challenging. Although the number of demonstrations in the \texttt{dense} setting is much larger than in the \texttt{sparse} setting, the contours of the effective range remain similar across both cases. This suggests more demonstrations near the center of the workspace do not significantly extend the effective range to more distant areas.

On the whole, the spatial generalization range of visuomotor policies can be roughly approximated by the union of the adjacent areas around the object configurations in the provided demonstrations. The extent of the adjacent regions is influenced by the fault tolerance level required for manipulation.

\vspace{0.2cm}\subsubsection{More Analysis on the Benchmarking Results in Fig.~\ref{fig:spatial_gen_benchmark}} 
\label{sec:appendix-empirical-benchmark}
On visuomotor policies, we find DP3 exhibits the highest spatial generalization capacity compared to all 2D-based counterparts. Additionally, models utilizing CLIP and DINOv2 representations achieve competitive results, significantly surpassing the train-from-scratch baseline. This highlights the value of pre-training on open-world vision tasks in enhancing spatial reasoning capabilities for robotic manipulation. Notably, while prior studies~\cite{chi2024universal,lin2024data,burns2023makes,zhu2024point} have emphasized the benefits of 3D and pre-trained encoders for visual generalization, our findings extend these insights to spatial generalization, offering a complementary perspective on encoder selection strategies.

On object randomization range, our experiments with \texttt{half} and \texttt{fixed} settings demonstrate that high precision requirements alone do not necessarily produce challenging tasks unless the object positions are fully randomized. This suggests that precision requirements and spatial randomization both contribute to the task difficulty.

On the number of demonstrations, while task performance generally improves with an increased number of demonstrations, the effect diminishes beyond a certain threshold. For instance, in the \texttt{full} workspace setting with DP3 as the policy, a $50$ demonstrations increase from $100$ to $150$ enhances the performance by $37\%$, but the increase from $150$ to $200$ only improves the performance by $6\%$.
This finding also highlights the inherent difficulty in achieving near-perfect success rates in robot learning systems.

\vspace{0.2cm}\subsubsection{The Precise-Peg-Insertion Task} 
\label{sec:appendix-empirical-task}
We construct a T-shaped peg, whose upper end has a cross-section of $6\,\textrm{cm} \times 6\,\textrm{cm}$, and the bottom end has a cross-section of $3\,\textrm{cm} \times 3\,\textrm{cm}$. The hole in the green socket has a cross-section of $4\,\textrm{cm} \times 4\,\textrm{cm}$. This shape enforces a strict fault tolerance of $1\,\textrm{cm}$ during both the picking and insertion stages, asking for millimeter-level precision. Both objects are randomized in a $40\,\mathrm{cm} \times 20\,\mathrm{cm}$ workspace in the \texttt{full} setting. The randomization range is halved into $20\,\mathrm{cm} \times 10\,\mathrm{cm}$ in the \texttt{half} setting.

\begin{figure}[h]
    \centering
    \vspace{-0.1cm}
    \includegraphics[width=\linewidth]{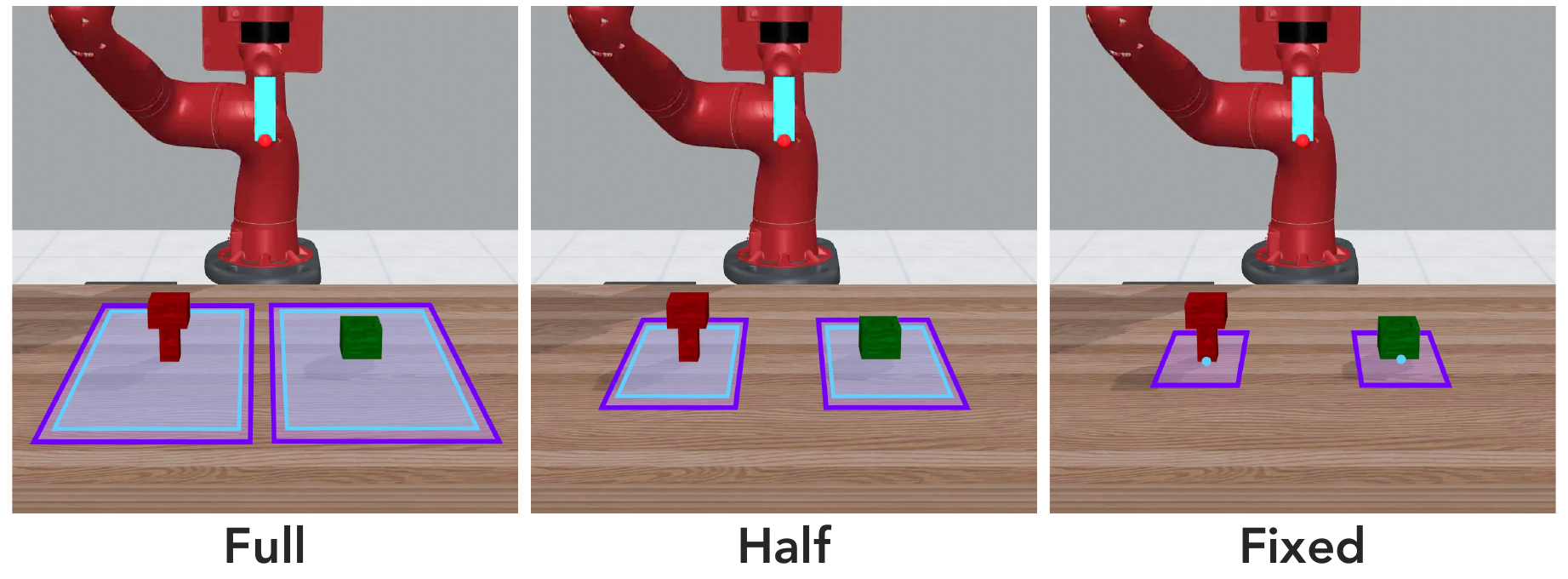}
    \caption{\textbf{The Precise-Peg-Insertion task.} A total of $3$ workspace sizes is considered. Purple and sky-blue rectangles mark the workspaces for demonstration and evaluation, respectively.}
    \label{fig:precise-peg-insertion}
\end{figure}

\begin{figure*}[t]
    \centering
    \includegraphics[width=\linewidth]{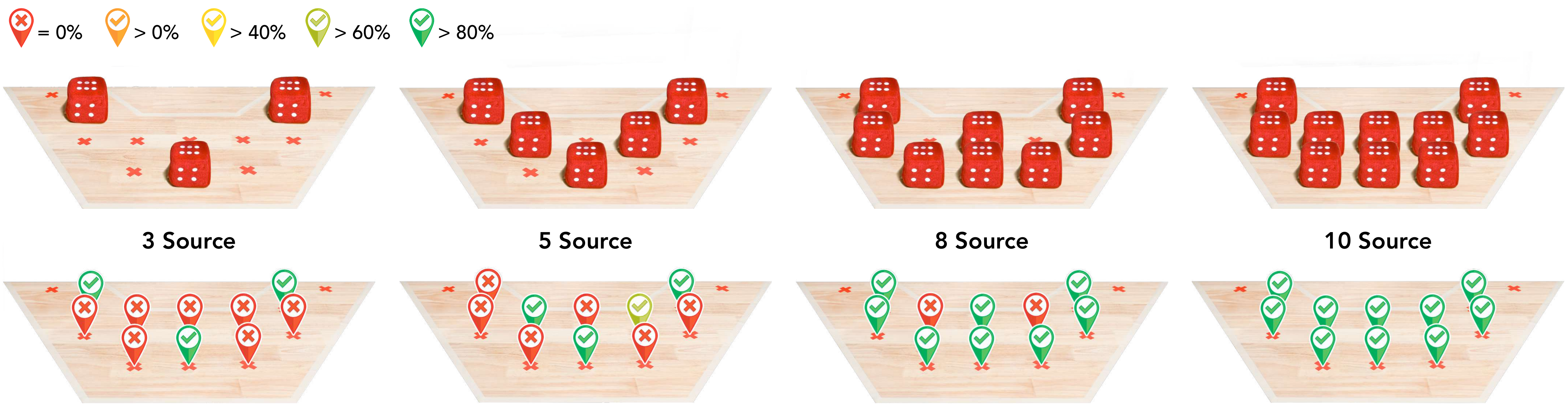}
    \caption{\textbf{Visualization of the policy performance trained on human-collected datasets.} (Upper row) The demonstrated configurations. (Bottom row) The spatial heatmaps with success rates averaged on $5$ trials.}
    \label{fig:source-demo-heatmap}
\end{figure*}

\subsection{Increased Human-Collected Demonstrations}
\label{sec:appendix-increase-source}

\begin{figure}
    \centering
    \includegraphics[width=0.95\linewidth]{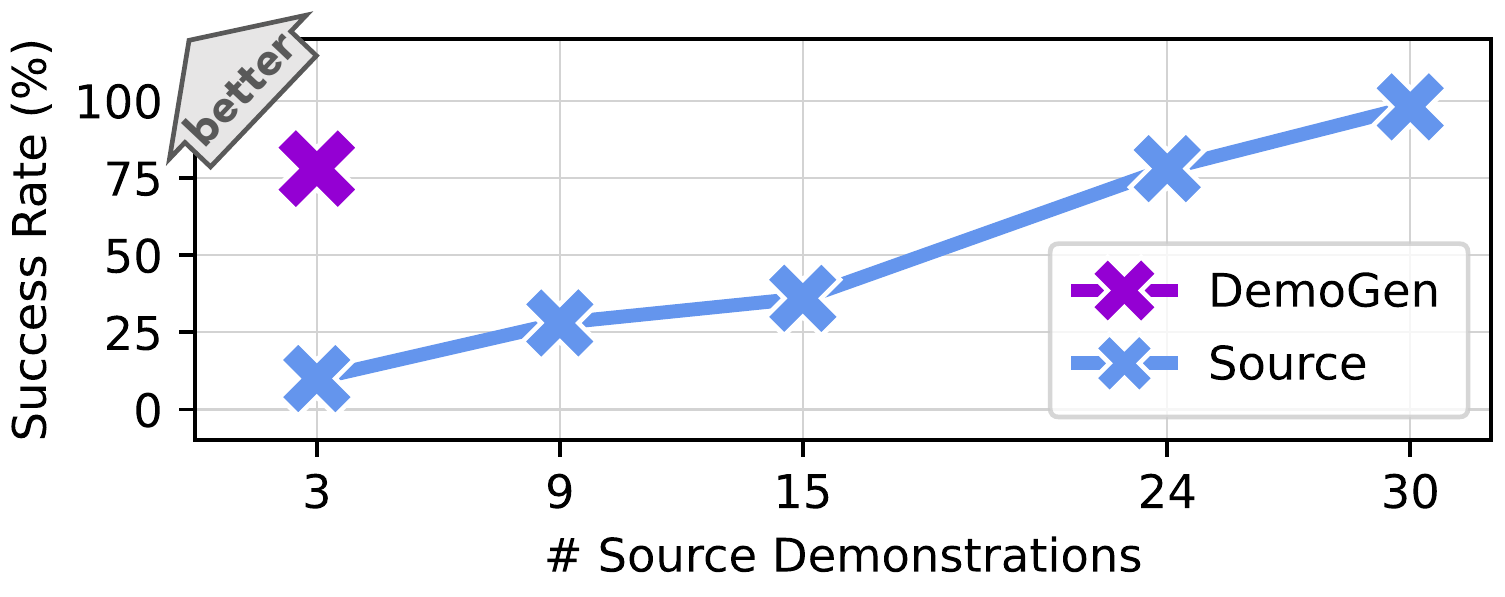}
    \caption{\textbf{Real-world comparison between \method-generated and human-collected datasets.} The \method-generated dataset is based on $3$ source demonstrations.}
    \label{fig:source-baseline}
\end{figure}

In Tab.~\ref{table:real-spatial-result}, we compare the \method-generated dataset against $3$ human-collected source demonstrations. 
In Fig.~\ref{fig:source-baseline}, we provide a reference on how the increase of source demonstrations leads to the enhancement of policy performance on the Dex-Cube task. 
To further understand the policy capacity enabled by human-collected demonstrations, we visualize the spatial heatmaps of human-collected datasets in Fig.~\ref{fig:source-demo-heatmap}. 
By comparing the demonstrated configurations and the spatial effective range of the resulting policies, we found the policy capacity is upper-bounded by the demonstrated configurations.
This is in line with the findings in the empirical study.

\subsection{Detailed Analysis of the Bimanual Humanoid Experiment}
\label{sec:appendix-humanoid}

The orientational augmentations share the same visual mismatch problem as translational augmentation. The policy performs as expected when the generated orientations are close to the orientation in the source demonstration. As the orientational difference increases, we observed the policy might react to the orientation in the current visual observation with actions for mismatched orientations.


Additionally, we found the spatial generalization problem persists in mobile manipulation scenarios. This is mainly due to the physical constraints of real-world environments, such as kitchen countertops or fruit stands, as demonstrated in our experiments, where terrain limitations prevent the base from approaching objects at arbitrary distances. Consequently, the base typically moves to a fixed point at a specific distance from the object, after which the robot conducts a standard non-mobile manipulation process at the fixed base position.

\subsection{Disturbance Resistance Experiments Details}
\label{sec:appendix-disturb}
\subsubsection{Evaluation Metrics}
\label{sec:appendix-disturb-metric}
The sauce coverage score is computed as follows. First, we distinguish between green background and red sauce in the HSV color space. The identified background is set to black, the sauce is set to red, and the rest which should be the uncovered crust is set to white. Second, due to the highlights on the sauce liquid, some small fragmented points of the sauce may be identified as the crust. To address this, we apply smoothing filtering followed by dilation and erosion, where the kernel size is $9\times 9$. Finally, the coverage is calculated as the ratio of red areas (sauce) over non-black areas (sauce + uncovered crust). 

\vspace{0.2cm}
\subsubsection{Raw Evaluation Results}
\label{sec:appendix-disturb-raw}
For quantitative evaluation, we perform $5$ repetitions for each of the $5$ disturbance directions, resulting in $25$ trials for both strategies. 

\end{appendix}

\begin{figure}
    \centering
    \includegraphics[width=\linewidth]{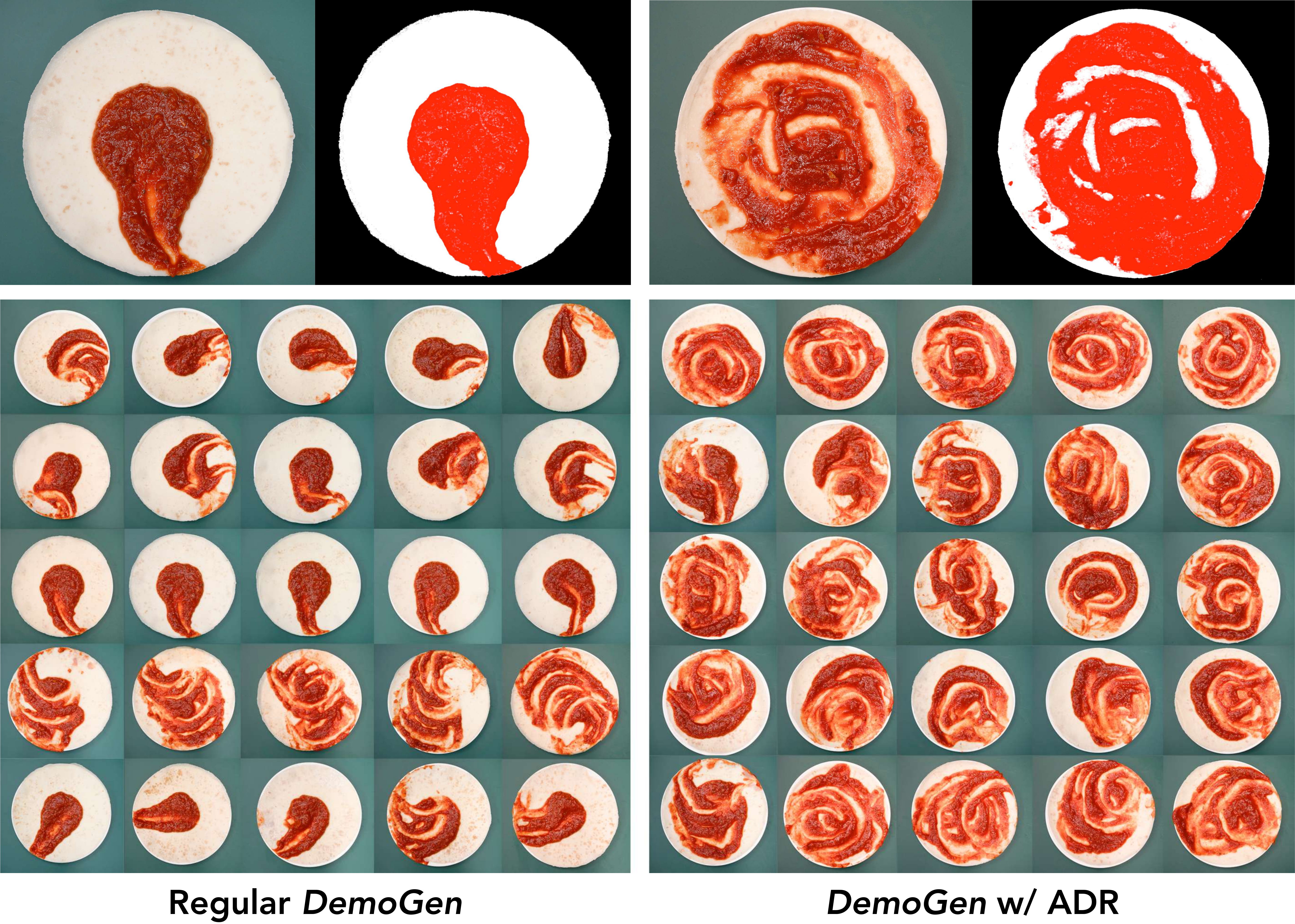}
    \caption{\textbf{Raw evaluation results in the Sauce-Spreading task.} (Top) Examples of the processing results for metric calculation. (Bottom) Compared with the regular \method, the policy trained with the ADR strategy better spreads the sauce to cover the crust under external disturbance.}
    \label{fig:sauce-raw-results}
    \vspace{-0.2cm}
\end{figure}

\subsection{Randomization Ranges for Simulated Tasks}
\label{sec:appendix-sim-range}

\input{tabs/sim-object-range}

In Fig.~\ref{fig:sim-tasks}, we illustrated the simulated tasks for the evaluation on spatial generalization. To strengthen the significance of spatial generalization, we enlarge the original object randomization ranges in the MetaWorld~\cite{yu2020metaworld} tasks. For demonstration generation, we select a slightly larger range than the evaluation workspace to avoid performance degradation near the workspace boundaries. The detailed workspace sizes are listed in Tab.~\ref{table:sim-range}.

\subsection{Task Descriptions for Real-World Tasks}
In Fig.~\ref{fig:real-spatial-tasks}, we illustrated the real-world tasks for the evaluation on spatial generalization. We describe these tasks in the text as follows, where we mark the verbs for  \textcolor{myorange}{motion} and \textcolor{myblue}{skill} actions in the corresponding colors.

\label{sec:appendix-task-real}
\begin{enumerate}
    \item \textbf{Spatula-Egg.} The gripper holds a spatula in hand. The robot maneuvers the spatula to first \textcolor{myorange}{move} toward the fried egg and then 1) \textcolor{myblue}{slide} beneath the egg, 2) \textcolor{myblue}{lift} the egg leveraging the contact with the plate's rim, 3) \textcolor{myblue}{carry} the egg and maintain stable suspension.
    \item \textbf{Flower-Vase.} The gripper \textcolor{myorange}{moves} toward the flower, \textcolor{myblue}{picks} it up, \textcolor{myorange}{reorients} it in the air while \textcolor{myorange}{transferring} toward the vase, and finally \textcolor{myblue}{inserts} it into the vase.
    \item \textbf{Mug-Rack.} The gripper \textcolor{myorange}{moves} toward the mug, \textcolor{myblue}{picks} it up, \textcolor{myorange}{reorients} it in the air while \textcolor{myorange}{transferring} toward the rack, and \textcolor{myblue}{hangs} it onto the rack.
    \item \textbf{Dex-Cube.} The dexterous hand \textcolor{myorange}{moves} toward the cube and \textcolor{myblue}{grasps} up the cube.
    \item \textbf{Dex-Rollup.} The dexterous hand \textcolor{myorange}{moves} toward a piece of plasticine and \textcolor{myblue}{wraps} it multiple times until it is fully coiled. The required times of the wrapping motion may vary due to the distinct plasticity of every hand-molded piece of plasticine.
    \item \textbf{Dex-Drill.} The dexterous hand \textcolor{myorange}{moves} toward the drill, \textcolor{myblue}{grasps} it up, \textcolor{myorange}{transfers} it toward the cube, and finally \textcolor{myblue}{touches} the cube with the drill. 
    \item \textbf{Dex-Coffee.} The dexterous hand \textcolor{myorange}{moves} toward the kettle, \textcolor{myblue}{grasps} it up, \textcolor{myorange}{transfers} it toward the coffee filter, and finally \textcolor{myblue}{pours} water into the filter.
\end{enumerate}

\subsection{Visualization of \method-Generated Trajectories}
\label{sec:appendix-vis-traj}
\begin{figure*}
    \centering
    \includegraphics[width=\linewidth]{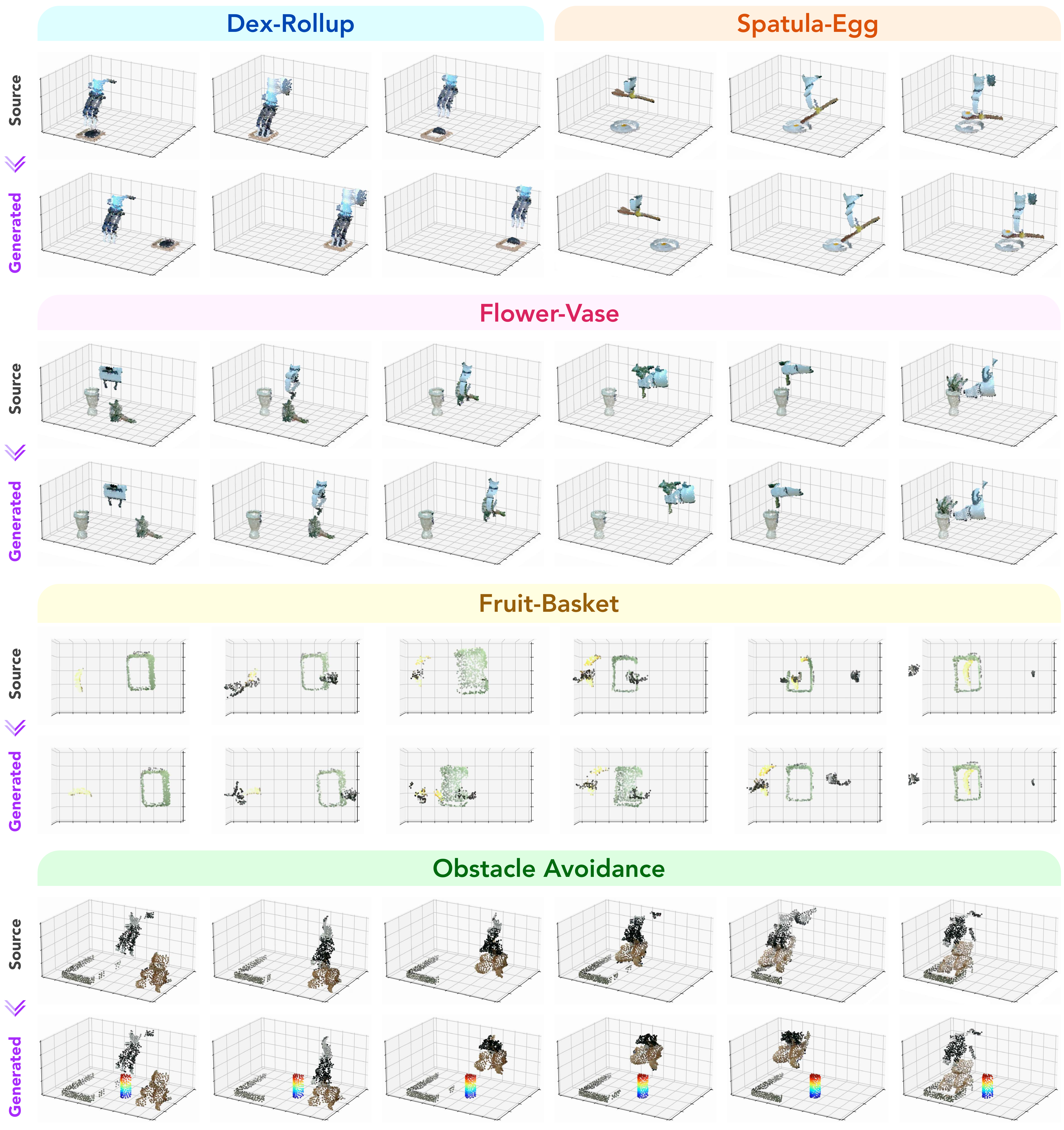}
    \caption{\textbf{More examples of the trajectories consisting of synthetic visual observations.} }
    \label{fig:traj-examples}
\end{figure*}

In Fig.~\ref{fig:method-traj}, we gave a concrete example of the trajectory of synthetic visual observations. We provide more examples in Fig.~\ref{fig:traj-examples} by showcasing the key frames of source and generated demonstrations.

%% file: tabs/sim-object-range.tex
\begin{table*}[t]
\centering
\caption{\textbf{Object randomization ranges in simulated tasks.} All the reported sizes have the units in centimeters.}
\label{table:sim-range}
\resizebox{1.0\textwidth}{!}{%
\begin{tabular}{l|cccccccc}
\toprule

& Pick-Cube & Button-Small & Drawer-Close & Faucet-Open & Handle-Press & Box-Lid & Stack-Cube & Assembly \\

\midrule

Object(s) & Cube & Button & Drawer & Faucet & Toaster & Box $\times$ Lid &  Red $\times$ Green & Pillar $\times$ Hole\\

Evaluation & $40 \!\times\! 40$ & $40 \!\times\! 40$ & $15 \!\times\! 15$ & $30 \!\times\! 30$ & $20 \!\times\! 30$ & $(2.5 \!\times\! 30)^2$ & $(15 \!\times\! 15)^2$ & $(10 \!\times\! 30)^2$ \\

\method & $48 \!\times\! 48$ & $48 \!\times\! 48$ & $20 \!\times\! 20$ & $40 \!\times\! 40$ & $25 \!\times\! 40$ & $(7.5 \!\times\! 40)^2$ & $(20 \!\times\! 20)^2$ & $(15 \!\times\! 40)^2$\\

\bottomrule
\end{tabular}}
\end{table*}